\documentclass{article}

\usepackage{times}

\usepackage{amsfonts}
\usepackage{amsmath}
\usepackage{bm}

\usepackage{graphicx}
\usepackage{subfigure}
\usepackage{multirow}

\usepackage{natbib}

\usepackage{hyperref}

\usepackage[accepted]{longform}

\usepackage{appendix}

\icmltitlerunning{Machine Theory of Mind}

\begin{document}

\twocolumn[
\icmltitle{Machine Theory of Mind}

\title{}

\vskip -0.5in
\author{
	\begin{tabular}[t]{c@{\extracolsep{4em}}c@{\extracolsep{4em}}c}
		{\bf Neil C. Rabinowitz\thanks{corresponding author}} & {\bf Frank Perbet} & {\bf H. Francis Song} \\
		DeepMind & DeepMind & DeepMind\\
		\texttt{ncr@google.com} & \texttt{fkp@google.com} & \texttt{songf@google.com} \\
	\\
	\\
	\\
		{\bf Chiyuan Zhang} & {\bf S. M. Ali Eslami} & {\bf Matthew Botvinick} \\
		Google Brain & DeepMind & DeepMind\\
		\texttt{chiyuan@google.com} & \texttt{aeslami@google.com} & \texttt{botvinick@google.com} \\
	\end{tabular}
}

\date{}
\maketitle
\thispagestyle{empty}
\vskip 0.1in
]


\newcommand\todo[1]{\textcolor{red}{#1}}

\newcommand{\leg}{\bf}

\printAffiliationsAndNotice{\icmlCorresponding}


\begin{abstract} 
Theory of mind \citep[ToM;][]{premack1978does} broadly refers to humans' ability to represent the mental states of others, including their desires, beliefs, and intentions. We propose to train a machine to build such models too. We design a Theory of Mind neural network~--~a {\it ToMnet}~--~which uses meta-learning to build models of the agents it encounters, from observations of their behaviour alone. Through this process, it acquires a strong prior model for agents' behaviour, as well as the ability to bootstrap to richer predictions about agents' characteristics and mental states using only a small number of behavioural observations. We apply the ToMnet to agents behaving in simple gridworld environments, showing that it learns to model random, algorithmic, and deep reinforcement learning agents from varied populations, and that it passes classic ToM tasks such as the ``Sally-Anne'' test \cite{wimmer1983beliefs, baron1985does} of recognising that others can hold false beliefs about the world. We argue that this system~--~which autonomously learns how to model other agents in its world~--~is an important step forward for developing multi-agent AI systems, for building intermediating technology for machine-human interaction, and for advancing the progress on interpretable AI.
\end{abstract} 



\section{Introduction}
\label{introduction}

For all the excitement surrounding deep learning and deep reinforcement learning at present, there is a concern from some quarters that our understanding of these systems is lagging behind. Neural networks are regularly described as opaque, uninterpretable black-boxes. Even if we have a complete description of their weights, it's hard to get a handle on what patterns they're exploiting, and where they might go wrong. As artificial agents enter the human world, the demand that we be able to understand them is growing louder.

Let us stop and ask: what does it actually mean to ``understand'' another agent? As humans, we face this challenge every day, as we engage with other humans whose latent characteristics, latent states, and computational processes are almost entirely inaccessible. Yet we function with remarkable adeptness. We can make predictions about strangers' future behaviour, and infer what information they have about the world; we plan our interactions with others, and establish efficient and effective communication.

A salient feature of these ``understandings'' of other agents is that they make little to no reference to the agents' true underlying structure. We do not typically attempt to estimate the activity of others' neurons, infer the connectivity of their prefrontal cortices, or plan interactions with a detailed approximation of the dynamics of others' hippocampal maps. A prominent argument from cognitive psychology is that our social reasoning instead relies on high-level {\it models} of other agents \cite{gopnik1992child}. These models engage abstractions which do not describe the detailed physical mechanisms underlying observed behaviour; instead, we represent the {\it mental states} of others, such as their desires, beliefs, and intentions. This ability is typically described as our Theory of Mind \cite{premack1978does}. While we may also, in some cases, leverage our own minds to simulate others' \citep[e.g.][]{gordon1986folk, gallese1998mirror}, our ultimate human understanding of other agents is not measured by a 1-1 correspondence between our models and the mechanistic ground truth, but instead by how much these models afford for tasks such as prediction and planning \cite{dennett1991two}.

In this paper, we take inspiration from human Theory of Mind, and seek to build a system which learns to model other agents. We describe this as a {\it Machine Theory of Mind}. Our goal is not to {\it assert} a generative model of agents' behaviour and an algorithm to invert it. Rather, we focus on the problem of how an observer could learn {\it autonomously} how to model other agents using limited data \cite{botvinick2017building}.  This distinguishes our work from previous literature, which has relied on hand-crafted models of agents as noisy-rational planners~--~e.g.\ using inverse RL \cite{ng2000algorithms, abbeel2004apprenticeship}, Bayesian inference \cite{lucas2014child, evans2016learning}, Bayesian Theory of Mind \cite{baker2011bayesian, jara2016naive, Baker2017} or game theory \cite{camerer2004cognitive, yoshida2008game, camerer2010behavioral, lanctot2017unified}. In contrast, we learn the agent models, and how to do inference on them, from scratch, via meta-learning.

Building a rich, flexible, and performant Machine Theory of Mind may well be a grand challenge for AI. We are not trying to solve all of this here. A main message of this paper is that many of the initial challenges of building a ToM can be cast as simple learning problems when they are formulated in the right way. Our work here is an exercise in figuring out these simple formulations.

There are many potential applications for this work. Learning rich models of others will improve decision-making in complex multi-agent tasks, especially where model-based planning and imagination are required \cite{hassabis2013imagine, hula2015monte, oliehoek2016concise}. Our work thus ties in to a rich history of opponent modelling \cite{brown1951iterative, albrecht2017autonomous}; within this context, we show how meta-learning could be used to furnish an agent with the ability to build flexible and sample-efficient models of others on the fly. Such models will be important for value alignment \cite{hadfield2016cooperative} and flexible cooperation \cite{nowak2006five, kleiman2016coordinate, barrett2017making, cao2018emergent}, and will likely be an ingredient in future machines' ethical decision making \cite{churchland1996neural}. They will also be highly useful for communication and pedagogy \cite{dragan2013legibility, fisac2017pragmatic, milli2017should}, and will thus likely play a key role in human-machine interaction. Exploring the conditions under which such abilities arise can also shed light on the origin of our human abilities \cite{carey2009origin}. Finally, such models will likely be crucial mediators of our human understanding of artificial agents.


Lastly, we are strongly motivated by the goals of making artificial agents human-interpretable. We attempt a novel approach here: rather than modifying agents architecturally to expose their internal states in a human-interpretable form, we seek to build intermediating systems which learn to reduce the dimensionality of the space of behaviour and re-present it in more digestible forms. In this respect, the pursuit of a Machine ToM is about building the missing interface between machines and human expectations \cite{cohen1981beyond}.

\subsection{Our approach}

We consider the challenge of building a Theory of Mind as essentially a meta-learning problem \cite{schmidhuber1996simple, thrun1998learning, hochreiter2001learning, vilalta2002perspective}. At test time, we want to be able to encounter a novel agent whom we have never met before, and already have a strong and rich prior about how they are going to behave. Moreover, as we see this agent act in the world, we wish to be able to collect data (i.e.\ form a posterior) about their latent characteristics and mental states that will enable us to improve our predictions about their future behaviour.

To do this, we formulate a meta-learning task. We construct an observer, who in each episode gets access to a set of behavioural traces of a novel agent. The observer's goal is to make predictions of the agent's future behaviour. Over the course of training, the observer should get better at rapidly forming predictions about new agents from limited data. This ``learning to learn'' about new agents is what we mean by meta-learning. Through this process, the observer should also learn an effective prior over the agents' behaviour that implicitly captures the commonalities between agents within the training population.

We introduce two concepts to describe components of this observer network and their functional role. We distinguish between a {\it general theory of mind} -- the learned weights of the network, which encapsulate predictions about the common behaviour of all agents in the training set -- and an {\it agent-specific theory of mind} -- the ``agent embedding'' formed from observations about a single agent at test time, which encapsulates what makes this agent's character and mental state distinct from others'. These correspond to a prior and posterior over agent behaviour.

This paper is structured as a sequence of experiments of increasing complexity on this Machine Theory of Mind network, which we call a {\it ToMnet}. These experiments showcase the idea of the ToMnet, exhibit its capabilities, and demonstrate its capacity to learn rich models of other agents incorporating canonical features of humans' Theory of Mind, such as the recognition of false beliefs.

Some of the experiments in this paper are directly inspired by the seminal work of Baker and colleagues in Bayesian Theory of Mind, such as the classic food-truck experiments \cite{baker2011bayesian, Baker2017}. We have not sought to directly replicate these experiments as the goals of this work differ. In particular, we do not immediately seek to explain human judgements in computational terms, but instead we emphasise machine learning, scalability, and autonomy. We leave the alignment to human judgements as future work. Our experiments should nevertheless generalise many of the constructions of these previous experiments.

Our contributions are as follows:
\begin{itemize}
    \item In Section~\ref{section:random-agents}, we show that for simple, random agents, the ToMnet learns to approximate Bayes-optimal hierarchical inference over agents' characteristics.
    \item In Section~\ref{section:inferring-goals}, we show that the ToMnet learns to infer the goals of algorithmic agents (effectively performing few-shot inverse reinforcement learning), as well as how they balance costs and rewards.
    \item In Section~\ref{section:trained-agents}, we show that the ToMnet learns to characterise different species of deep reinforcement learning agents, capturing the essential factors of variations across the population, and forming abstract embeddings of these agents. We also show that the ToMnet can discover new abstractions about the space of behaviour.
    \item In Section~\ref{section:implicit-beliefs}, we show that when the ToMnet is trained on deep RL agents acting in POMDPs, it implicitly learns that these agents can hold false beliefs about the world, a core component of humans' Theory of Mind.
    \item In Section~\ref{section:explicit-beliefs}, we show that the ToMnet can be trained to predict agents' belief states as well, revealing agents' false beliefs explicitly. We also show that the ToMnet can infer what different agents are able to see, and what they therefore will tend to believe, from their behaviour alone. 
\end{itemize}


\section{Model}
\label{model}

\subsection{The tasks}
\label{section:tasks}

\newcommand{\Agent}{\mathcal{A}}
\newcommand{\MDP}{\mathcal{M}}
\newcommand{\Npast}{N_{\mathrm{past}}}

Here we describe the formalisation of the task. We assume we have a family of partially observable Markov decision processes (POMDPs) $\MDP = \bigcup_{j} \MDP_j$. Unlike the standard formalism, we associate the reward functions, discount factors, and conditional observation functions with the agents rather than with the POMDPs. For example, a POMDP could be a gridworld with a particular arrangement of walls and objects; different agents, when placed in the same POMDP, might receive different rewards for reaching these objects, and be able to see different amounts of their local surroundings. The POMDPs are thus tuples of state spaces $S_j$, action spaces $A_j$, and transition probabilities $T_j$ only, i.e.\ $\MDP_j = (S_j, A_j, T_j)$. In this work, we only consider single-agent POMDPs, though the extension to the multi-agent case is simple. When agents have full observability, we use the terms MDP and POMDP interchangeably. We write the joint state space over all POMDPs as $S = \bigcup_j S_j$.

Separately, we assume we have a family of agents $\Agent = \bigcup_i \Agent_i$, with corresponding observation spaces $\Omega_i$, conditional observation functions $\omega_i(\cdot): S \rightarrow \Omega_i$, reward functions $R_i$, discount factors $\gamma_i$, and resulting policies $\pi_i$, i.e.\ $\Agent_i = (\Omega_i, \omega_i, R_i, \gamma_i, \pi_i)$. These policies might be stochastic (as in Section~\ref{section:random-agents}), algorithmic (as in Section~\ref{section:inferring-goals}), or learned (as in Sections~\ref{section:trained-agents}--\ref{section:explicit-beliefs}). We do not assume that the agents' policies $\pi_i$ are optimal for their respective tasks. The agents may be stateful -- i.e.\ with policies parameterised as $\pi_i(\cdot | \omega_i(s_t), h_t)$ where $h_t$ is the agent's (Markov) hidden state -- though we assume agents' hidden states do not carry over between episodes. 

In turn, we consider an observer who makes potentially partial and/or noisy observations of agents' trajectories, via a state-observation function $\omega^{(obs)}(\cdot): S \rightarrow \Omega^{(obs)}$, and an action-observation function $\alpha^{(obs)}(\cdot): A \rightarrow A^{(obs)}$. Thus, if agent $\Agent_i$ follows its policy $\pi_i$ on POMDP $\MDP_j$ and produces trajectory $\tau_{ij} = \{(s_t, a_t)\}_{t=0}^{T}$, the observer would see $\tau_{ij}^{(obs)} = \{(x_t^{(obs)}, a_t^{(obs)})\}_{t=0}^{T}$, where $x_t^{(obs)} = \omega^{(obs)}(s_t)$ and $a_t^{(obs)} = \alpha^{(obs)}(a_t)$. For all experiments we pursue here, we set $\omega^{(obs)}(\cdot)$ and $\alpha^{(obs)}(\cdot)$ as identity functions, so that the observer has unrestricted access to the MDP state and overt actions taken by the agents; the observer does not, however, have access to the agents' parameters, reward functions, policies, or identifiers.

We set up the meta-learning problem as follows. ToMnet training involves a series of encounters with individual agents, together with a query for which the ToMnet has to make a set of predictions. More precisely, the observer sees a set of full or partial ``past episodes'', wherein a single, unlabelled agent, $\Agent_i$, produces trajectories, $\{\tau_{ij}\}_{j=1}^{\Npast}$, as it executes its policy within the respective POMDPs, $\MDP_j$. Generally, we allow $\Npast$ to vary, sometimes even setting it to zero. The task for the observer is to predict the agent's behaviour (e.g.\ atomic actions) and potentially its latent states (e.g.\ beliefs) on a ``current episode'' as it acts within POMDP $\MDP_k$. The observer may be seeded with a partial trajectory in $\MDP_k$ up to time $t$.

The observer must learn to predict the behaviour of {\it many} agents, whose rewards, parameterisations, and policies may vary considerably; in this respect, the problem resembles the one-shot imitation learning setup recently introduced in \citet{Duan2017} and \citet{wang2017robust}. However, the problem statement differs from imitation learning in several crucial ways. First, the observer need not be able to execute the behaviours itself: the behavioural predictions may take the form of atomic actions, options, trajectory statistics, or goals or subgoals. The objective here is not to imitate, but instead to form predictions and abstractions that will be useful for a range of other tasks. Second, there is an informational asymmetry, where the ``teacher'' (i.e.\ the agent $\Agent_i$) may conceivably know {\it less} about the environment state $s_t$ than the ``student'' (i.e.\ the observer), and it may carry systematic biases; its policy, $\pi_i$, may therefore be far from optimal. As a result, the observer may need to factor in the likely knowledge state of the agent and its cognitive limitations when making behavioural predictions. Finally, as a ToM needs to operate online while observing a new agent, we place a high premium on the speed of inference. Rather than using the computationally costly algorithms of classical inverse reinforcement learning \citep[e.g.][]{ng2000algorithms, ramachandran2007bayesian, ziebart2008maximum, boularias2011relative}, or Bayesian ToM \citep[e.g.][]{baker2011bayesian, nakahashi2016modeling, Baker2017}, we drive the ToMnet to amortise its inference through neural networks \citep[as in][]{kingma2013auto, rezende2014stochastic, ho2016generative, Duan2017, wang2017robust}.

\subsection{The architecture}

\newcommand{\echar}{e_{\mathrm{char}}}
\newcommand{\echari}{e_{\mathrm{char,i}}}
\newcommand{\echarij}{e_{\mathrm{char},ij}}
\newcommand{\emental}{e_{\mathrm{mental}}}
\newcommand{\ementali}{e_{\mathrm{mental},i}}
\newcommand{\ementalij}{e_{\mathrm{mental},ij}}

\begin{figure}[!t]
\begin{center}
\centerline{\includegraphics[width=\columnwidth]{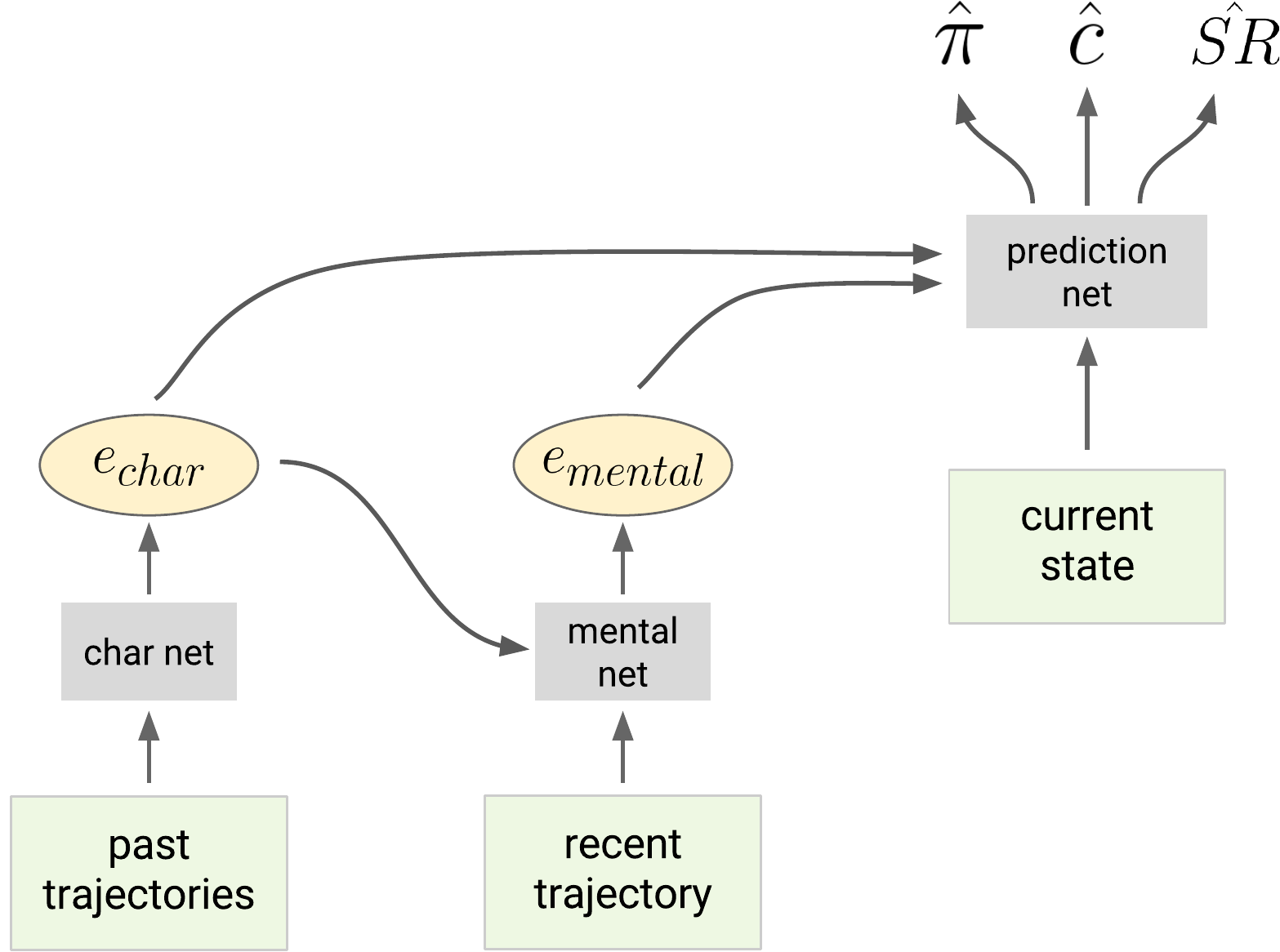}}
	\caption{{\leg ToMnet architecture.} The {\it character net} parses an agent's past trajectories from a set of POMDPs to form a character embedding, $\echar$. The {\it mental state net} parses the agent's trajectory on the current episode, to form an embedding of its mental state, $\emental$. These embeddings are fed into the {\it prediction net}, which is then queried with a current state. This outputs predictions about future behaviour, such as next-step action probabilities ($\hat{\pi}$), probabilities of whether certain objects will be consumed ($\hat{c}$), and predicted successor representations \citep[$\widehat{SR}$;][]{dayan1993improving}.}
\label{fig:model}
\end{center}
\vskip -0.2in
\end{figure}

To solve these tasks, we designed the {\it ToMnet} architecture shown in Fig~\ref{fig:model}. The ToMnet is composed of three modules: a {\it character net}, a {\it mental state net}, and a {\it prediction net}.

The goal of the character net is to {\it characterise} the presented agent, by parsing observed past episode trajectories, $\{\tau_{ij}^{(obs)}\}_{j=1}^{\Npast}$, into a character embedding, $\echari$. Here we choose to parse each past episode independently using a learned neural net, $f_{\theta}$, as $\echarij = f_{\theta} \left( \tau_{ij}^{(obs)} \right)$, and sum these to form the embedding $\echari = \sum_{j=1}^{\Npast} \echarij$.

The goal of the mental state net is to {\it mentalise} about the presented agent during the current episode \citep[i.e.\ infer its mental state;][]{dennett1973intentional, frith2006neural}, by parsing the current episode trajectory, $\tau_{ik}^{(obs)}$, up to time $t-1$ into a mental state embedding, $\ementali$, using a learned neural net, $g_{\phi}$. This takes the form $\ementali = g_{\phi} \left( [\tau_{ij}^{(obs)}]_{0:t-1}, \echari \right)$. For brevity, we drop the agent subscript, $i$.

Lastly, the goal of the prediction net is to leverage the character and mental state embeddings to predict subsequent behaviour of the agent. For example, next-step action prediction takes the form of estimating the given agent's policy with $\hat{\pi}(\cdot | x_t^{(obs)}, \echar, \emental)$. We also predict other behavioural quantities, described below. We use a shared torso and separate heads for the different prediction targets. Precise details of the architecture, loss, and hyperparameters for each experiment are given in Appendix~\ref{appendix:tomnet}. We train the whole ToMnet end-to-end.

\subsection{Agents and environments}

We deploy the ToMnet to model agents belonging to a number of different ``species'' of agent. In Section~\ref{section:random-agents}, we consider species of agents with random policies. In Section~\ref{section:inferring-goals}, we consider species of agents with full observability over MDPs, which plan using value iteration. In Sections~\ref{section:trained-agents}~--~\ref{section:explicit-beliefs}, we consider species of agents with different kinds of partial observability (i.e.\ different functions $\omega_i(\cdot)$), with policies parameterised by feed-forward nets or LSTMs. We trained these agents using a version of the UNREAL deep RL framework \cite{jaderberg2017reinforcement}, modified to include an auxiliary belief task of estimating the locations of objects within the MDP. Crucially, we did not change the core architecture or algorithm of the ToMnet observer to match the structure of the species, only the ToMnet's capacity.

The POMDPs we consider here are all gridworlds with a common action space (up/down/left/right/stay), deterministic dynamics, and a set of consumable objects, as described in the respective sections and in Appendix~\ref{appendix:gridworlds}. We experimented with these POMDPs due to their simplicity and ease of control; our constructions should generalise to richer domains too. We parameterically generate individual $\MDP_j$ by randomly sampling wall, object, and initial agent locations.


\section{Experiments}
\label{experiments}

\newcommand{\Species}{\mathcal{S}}
\newcommand{\actprob}{\bm{\pi_i}}

\subsection{Random agents}
\label{section:random-agents}

\begin{figure}[!t]
\begin{center}
\centerline{\includegraphics[width=\columnwidth]{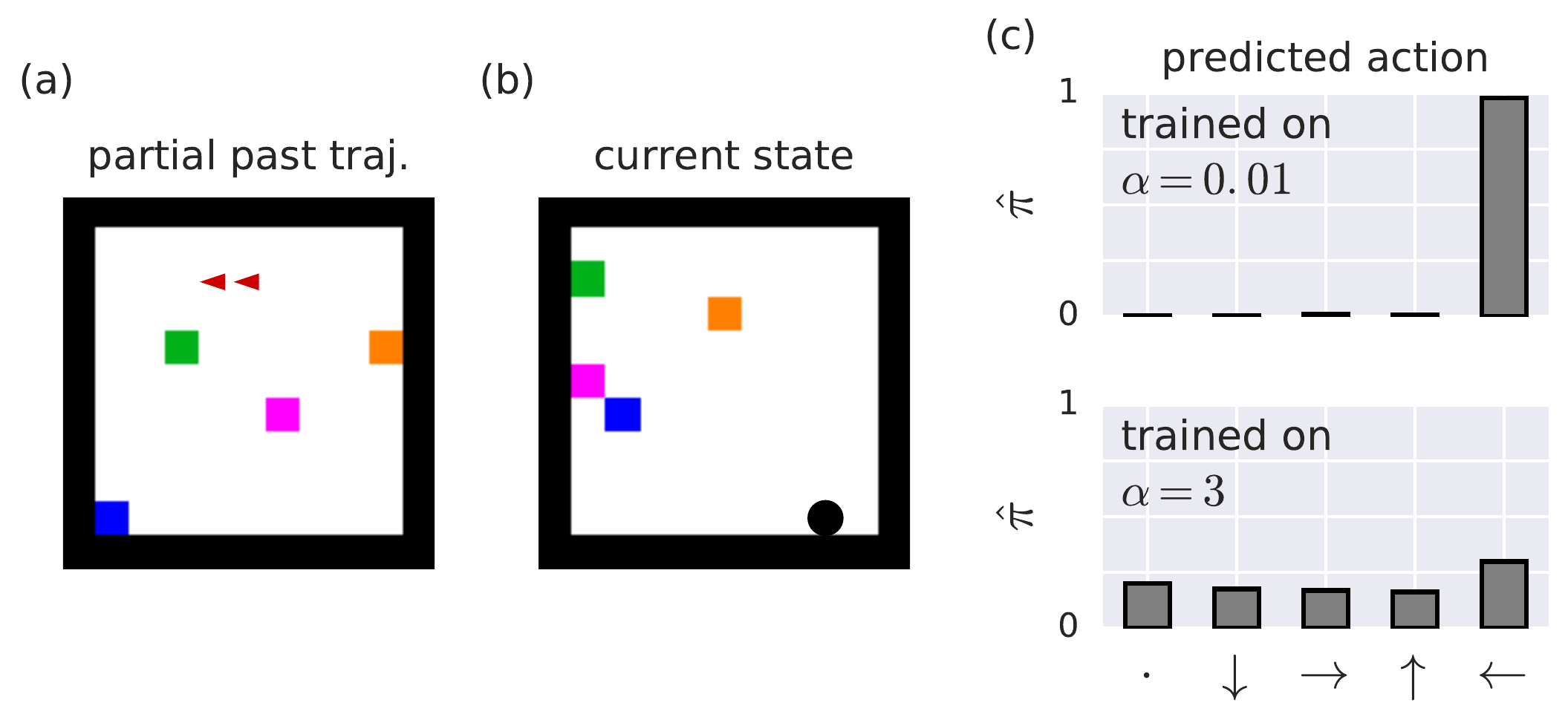}}
	\caption{{\leg Example gridworld in which a random agent acts.} {\leg (a)} Example past episode. Coloured squares indicate objects. Red arrows indicate the positions and actions taken by the agent. {\leg (b)} Example query: a state from a new MDP. Black dot indicates agent position. {\leg (c)} Predictions for the next action taken by the agent shown in (a) in query state (b). Top: prediction from ToMnet trained on agents with near-deterministic policies. Bottom: prediction from ToMnet trained on agents with more stochastic policies.}
\label{fig:random-agents-1}
\end{center}
\vskip -0.2in
\end{figure}

To demonstrate its essential workings, we tested the ToMnet observer on a simple but illustrative toy problem. We created a number of different {\it species} of random agents, sampled agents from them, and generated behavioural traces on a distribution of random $11 \times 11$ gridworlds (e.g.\ Fig~\ref{fig:random-agents-1}a). Each agent had a stochastic policy defined by a fixed vector of action probabilities $\pi_i(\cdot) = \actprob$. We defined different species based on how sparse its agents' policies were: within a species $\Species(\alpha)$, each $\actprob$ was drawn from a Dirichlet distribution with concentration parameter $\alpha$. At one extreme, we created a species of agents with near-deterministic policies by drawing $\actprob \sim Dir(\alpha=0.01)$; here a single agent might overwhelmingly prefer to always move left, and another to always move up. At the other extreme, we created a species of agent with far more stochastic policies, by drawing $\actprob \sim Dir(\alpha = 3)$.

Next, we trained different ToMnet observers each on a single species of agent. For each $\alpha$, we formed a training set by sampling 1000 agents from $\Species(\alpha)$, and for each agent, generating behavioural traces on randomly-generated POMDPs. We then trained a ToMnet to observe how randomly-sampled agents $\Agent_i \sim \Species(\alpha)$ behave on a variable number of past episodes ($\Npast \sim U\{0, 10\}$; for simplicity, limiting the length of each past trajectory to a single observation/action pair) and to use this information to predict the initial action that each agent $\Agent_i$ would take in a new POMDP, $\MDP_k$ (e.g.\ Fig~\ref{fig:random-agents-1}b-c). We omitted the mental net for this task.

\begin{figure}[!t]
\begin{center}
\centerline{\includegraphics[width=\columnwidth]{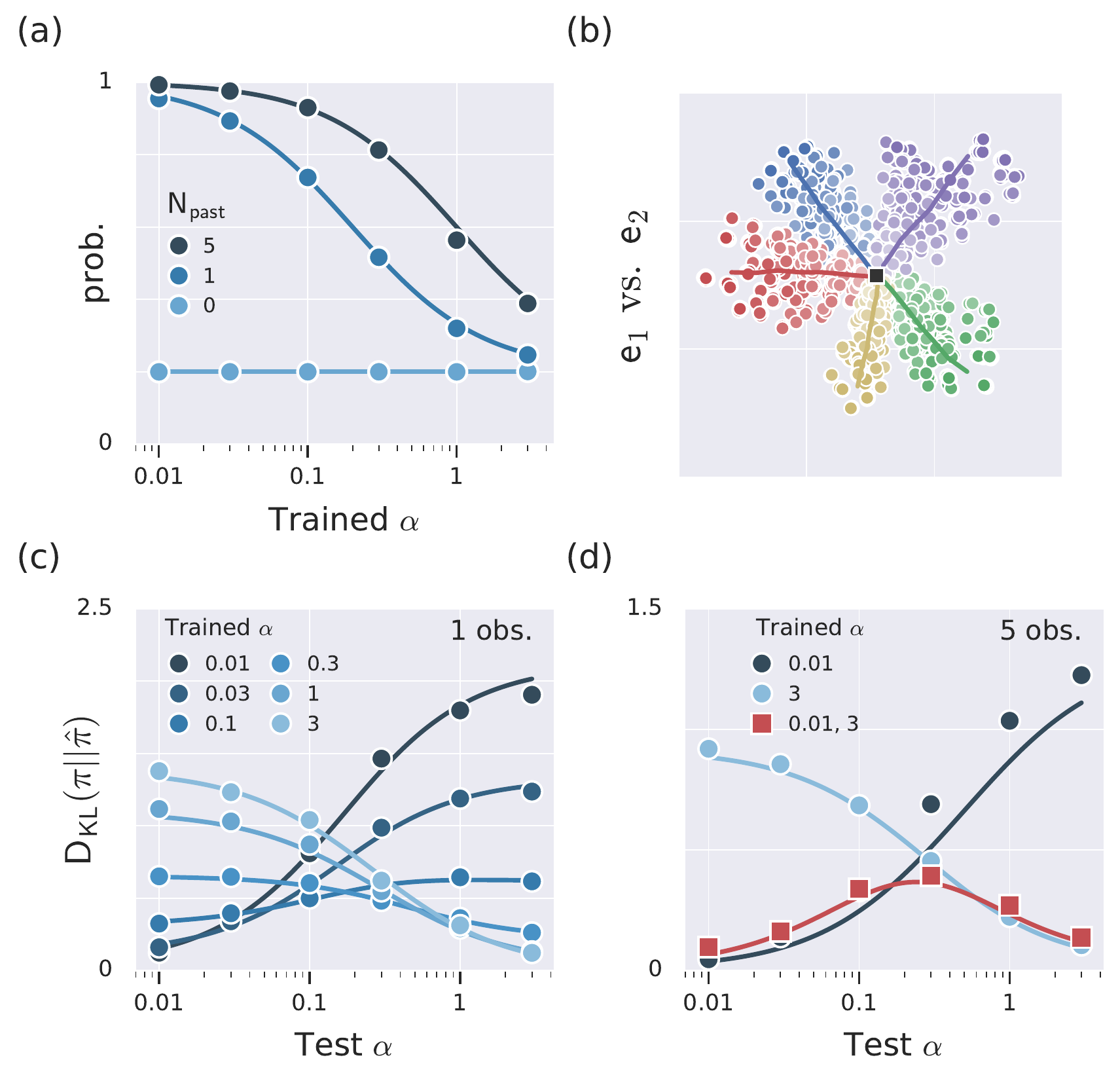}}
	\caption{{\leg ToMnet trained on random agents.} {\leg (a)} Likelihood of agent's true actions under the ToMnet's predictions, given that the ToMnet has been trained on species $\Species(\alpha)$. Priors are shown in light blue, and posteriors after observing that agent perform just that same action in $\Npast = 1$ or $5$ past episodes in darker blue. Dots are data from the ToMnet; solid lines are from the analytic Bayes-optimal posteriors specialised to the respective $\Species(\alpha)$. {\leg (b)} Character embeddings $\echar \in \mathbb{R}^2$ of different agents. Dots are coloured by which action was observed to occur most during $\Npast=10$ past episodes, and are darker the higher that count. {\leg (c)} Average KL-divergence between agents' true and predicted policies when the ToMnet is trained on agents from one species, $\Species(\alpha)$, but tested on agents from a different species $\Species(\alpha^\prime)$. Dots show values from the ToMnet; lines show analytic expected KLs when using analytic Bayes-optimal inference as in (a). Values calculated for $\Npast=1$. The ToMnet thus learns an effective prior for the species it is trained on. {\leg (d)} Same, but including a ToMnet trained on a mixture of species (with $\Npast=5$). The ToMnet here implicitly learns to perform hierarchical inference.}
\label{fig:random-agents-2}
\end{center}
\vskip -0.2in
\end{figure}

When the ToMnet observer is trained on a species $\Species(\alpha)$, it learns how to approximate Bayes-optimal, online inference about agents' policies $\pi_i(\cdot) = \actprob \sim Dir(\alpha)$. Fig~\ref{fig:random-agents-2}a shows how the ToMnet's estimates of action probability increase with the number of past observations of that action, and how training the ToMnet on species with lower $\alpha$ makes it apply priors that the policies are indeed sparser. We can also see how the ToMnet specialises to a given species by testing it on agents from different species (Fig~\ref{fig:random-agents-2}c): the ToMnet makes better predictions about novel agents drawn from the species which it was trained on. Moreover, the ToMnet easily learns how to predict behaviour from mixtures of species (Fig~\ref{fig:random-agents-2}d): when trained jointly on species with highly deterministic ($\alpha = 0.01$) and stochastic ($\alpha = 3$) policies, it implicitly learns to expect this bimodality in the policy distribution, and specialises its inference accordingly. We note that it is not learning about two {\it agents}, but rather two {\it species} of agents, which each span a spectrum of individual parameters.

There should be nothing surprising about seeing the ToMnet learn to approximate Bayes-optimal online inference; this should be expected given more general results about inference and meta-learning with neural networks \cite{mackay1995probable, finn2017meta}. Our point here is that a very first step in reasoning about other agents is an inference problem. The ToMnet is just an engine for learning to do inference and prediction on other agents.

The ToMnet does expose an agent embedding space which we can explore. In Fig~\ref{fig:random-agents-2}b, we show the values of $\echar$ produced by a ToMnet with a 2D embedding space. We note that the Bayes-optimal estimate of an agent's policy is a Dirichlet posterior, which depends only on $\alpha$ (which is fixed for the species) and on the observed action count (a 5-dim vector). We see a similar solution reflected in the ToMnet's $\echar$ embedding space, wherein agents are segregated along canonical directions by their empirical action counts.

In summary, without any changes to its architecture, a ToMnet learns a {\it general theory of mind} that is specialised for the distribution of agents it encounters in the world, and estimates an {\it agent-specific theory of mind} online for each individual agent that captures the sufficient statistics of its behaviour.


\subsection{Inferring goal-directed behaviour}
\label{section:inferring-goals}

An elementary component of humans' theory of other agents is an assumption that agents' behaviour is {\it goal-directed}. There is a wealth of evidence showing that this is a core component of our model from early infancy \cite{Gergely1995, Woodward1998, Woodward1999, Buresh2007}, and intelligent animals such as apes and corvids have been shown to have similar expectations about their conspecifics \cite{call2008does, ostojic2013evidence}. Inferring the desires of others also takes a central role in machine learning in imitation learning, most notably in inverse RL \cite{ng2000algorithms, abbeel2004apprenticeship}.

We demonstrate here how the ToMnet observer learns how to infer the goals of reward-seeking agents. We defined species of agents who acted within gridworlds with full observability (Fig~2a). Each gridworld was $11\times11$ in size, had randomly-sampled walls, and contained four different objects placed in random locations. Consuming an object yielded a reward for the agent and caused the episode to terminate. Each agent, $\Agent_i$, had a unique, fixed reward function, such that it received reward $r_{i,a} \in (0, 1)$ when it consumed object $a$; the vectors $\mathbf{r_i}$ were sampled from a Dirichlet distribution with concentration parameter $\alpha=0.01$. Agents also received a negative reward of $-0.01$ for every move taken, and a penalty of $0.05$ for walking into walls. In turn, the agents planned their behaviour through value iteration, and hence had optimal policies $\pi_i^*$ with respect to their own reward functions.

We trained the ToMnet to observe behaviour of these agents in randomly-sampled ``past'' MDPs, and to use this to predict the agents' behaviour in a ``current'' MDP. We detail three experiments below; these explore the range of capabilities of the ToMnet in this domain.

\begin{figure}[!t]
\begin{center}
\centerline{\includegraphics[width=\columnwidth]{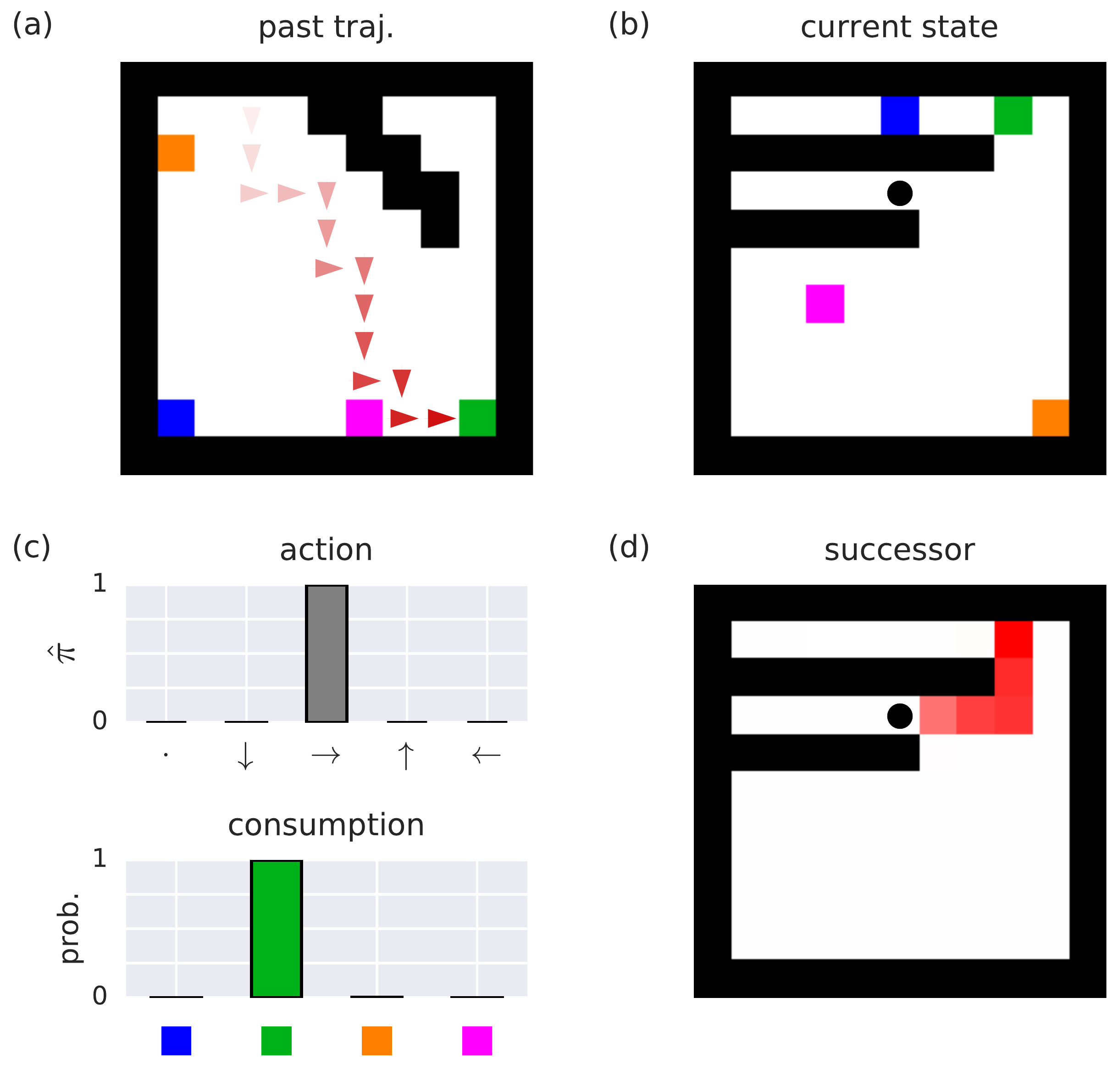}}
	\caption{{\leg ToMnet on goal-driven agents.} {\leg (a)} Past trajectory of an example agent. Coloured squares indicate the four objects. Red arrows indicate the position and action taken by the agent. {\leg (b)} Example query: a state from a new MDP. Black dot indicates agent position. {\leg (c)} ToMnet's prediction for the agent's next action (top) and object consumed at the end of the episode (bottom) for the query MDP in (b), given the past observation in (a). {\leg (d)} ToMnet's prediction of the successor representation (SR) for query (b), using discount $\gamma = 0.9$. Darker shading indicates higher expected discounted state occupancy.}
\label{fig:goal-driven-1}
\end{center}
\vskip -0.2in
\end{figure}

First, we provided the ToMnet with a full trajectory of an agent on a single past MDP (Fig~\ref{fig:goal-driven-1}a). In turn, we queried the ToMnet with the initial state of a current MDP (Fig~\ref{fig:goal-driven-1}b) and asked for a set of predictions: the next action the agent would take (Fig~\ref{fig:goal-driven-1}c top), what object the agent would consume by the end of the episode (Fig~\ref{fig:goal-driven-1}c bottom), and a set of statistics about the agent's trajectory in the current MDP, the successor representation \citep[SR; the expected discounted state occupancy;][Fig~\ref{fig:goal-driven-1}]{dayan1993improving}. The ToMnet's predictions qualitatively matched the agents' true behaviours.

\begin{figure}[!t]
\begin{center}
\centerline{\includegraphics[width=\columnwidth]{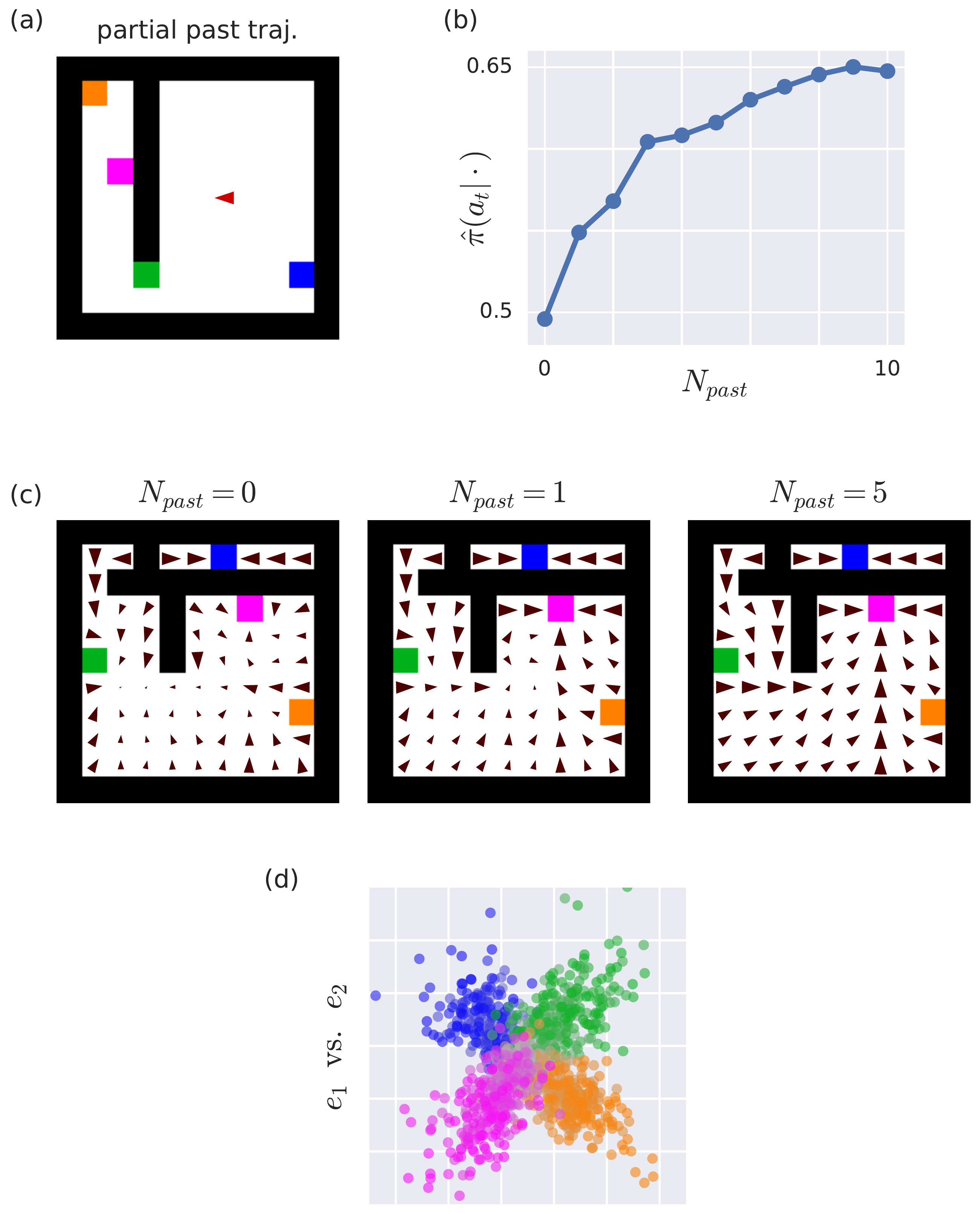}}
	\caption{{\leg ToMnet on goal-driven agents, continued.} {\leg (a)} This ToMnet sees only snapshots of single observation/action pairs (red arrow) from a variable number of past episodes (one shown here). {\leg (b)} Increasing $\Npast$ leads to better predictions; here we show the average posterior probability assigned to the true action. Even when $\Npast=0$, the action probability is greater than chance, since all agents in the species have similar policies in some regions of the state space. {\leg (c)} Predicted policy for different initial agent locations in a query MDP, for different numbers of past observations. Arrows show resultant vectors for the predicted policies, i.e. $\sum_k \mathbf{a_k} \cdot \hat{\pi}(\mathbf{a_k} | x, \echar)$. When $\Npast = 0$, the ToMnet has no information about the agent's preferred object, so the predicted policy exhibits no net object preference. When $\Npast>0$, the ToMnet infers a preference for the pink object. When the agent is stuck in the top right chamber, the ToMnet predicts that it will always consume the blue object, as this terminates the episode as soon as possible, avoiding a costly penalty. {\leg (d)} 2D embedding space of the ToMnet, showing values of $\echar$ from 100 different agents. Agents are colour-coded by their ground-truth preferred objects; saturation increases with $\Npast$, with the grey dots in the centre denoting agents with $\Npast=0$.}
\label{fig:goal-driven-2}
\end{center}
\vskip -0.2in
\end{figure}

Second, as a more challenging task, we trained a ToMnet to observe only partial trajectories of the agent's past behaviour. We conditioned the ToMnet on single observation-action pairs from a small number of past MDPs ($\Npast \sim \mathcal{U}\{0, 10\}$; e.g.\ Fig~\ref{fig:goal-driven-2}a). As expected, increasing the number of past observations of an agent improved the ToMnet's ability to predict its behaviour on a new MDP (Fig~\ref{fig:goal-driven-2}b), but even in the absence of any past observations, the ToMnet had a strong prior for the reasonable behaviour that would be expected of any agent within the species, such as movement away from the corners, or consumption of the only accessible object (Fig~\ref{fig:goal-driven-2}c). 

We note that unlike the approach of inverse RL, the ToMnet is not constrained to explicitly infer the agents' reward functions in service of its predictions. Nevertheless, in this simple task, using a 2-dimensional character embedding space renders this information immediately legible (Fig~\ref{fig:goal-driven-2}d). This is also true when the only behavioural prediction is next-step action.

\begin{figure}[!t]
\begin{center}
\centerline{\includegraphics[width=\columnwidth]{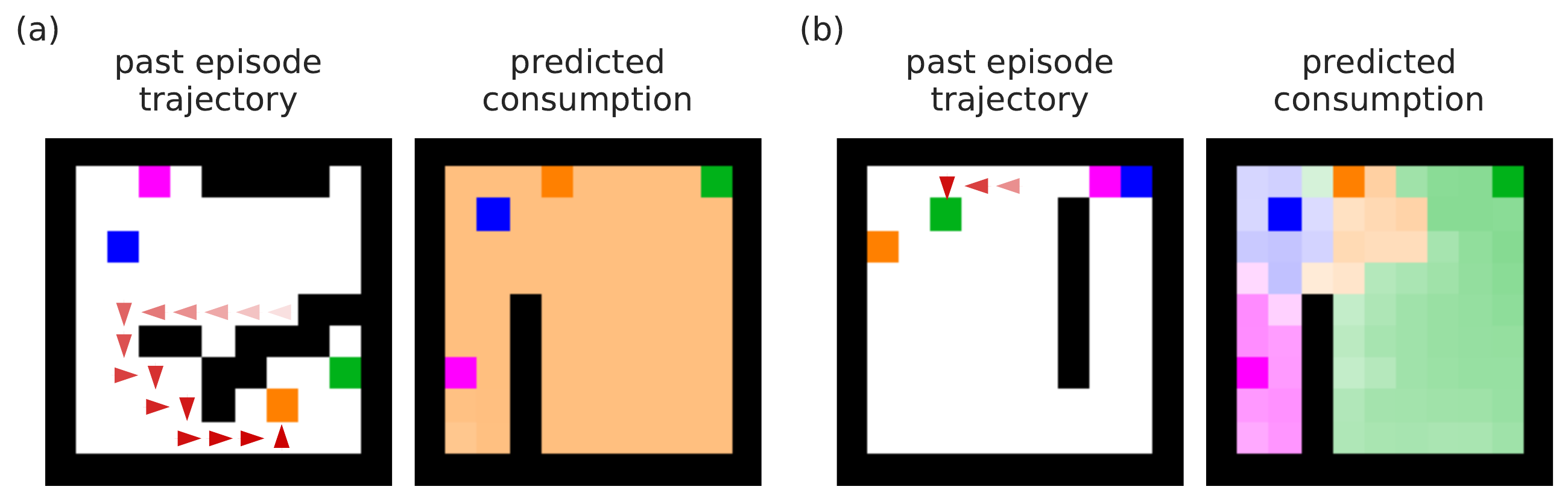}}
	\caption{{\leg ToMnet on greedy agents.} Left: a single past trajectory. Right: ToMnet predictions on a query MDP. Light shaded regions indicate ToMnet's prediction for the most probable object the agent will consume by the end of the episode, given that the agent is currently in that location. {\leg (a)} After seeing the agent take a long path to the orange object, the ToMnet predicts it will try to consume the orange object on the query MDP, no matter its current location. {\leg (b)} After seeing the agent take the shortest path to the green object, the ToMnet predicts it will generally consume a nearby object on the query MDP.}
\label{fig:goal-driven-3}
\end{center}
\vskip -0.2in
\end{figure}

Finally, we added more diversity to the agent species by applying a very high move cost ($0.5$) to 20\% of the agents; these agents therefore generally sought the closest object. We trained a ToMnet to observe a small number of full trajectories ($\Npast \sim \mathcal{U}\{0, 5\}$) of randomly-selected agents before making its behavioural prediction. The ToMnet learned to infer from even a single trajectory which subspecies of agent it was observing: if the agent went out of its way to consume a distant object on a past episode, then the ToMnet inferred a strong posterior that it would do so in a new episode from any starting position (Fig~\ref{fig:goal-driven-3}a); if the agent sought the closest object in a past episode, then the ToMnet was more cautious about whether it would seek the same object again on a new episode, deferring instead to a prediction that the agent would act greedily again (Fig~\ref{fig:goal-driven-3}b). This inference resembles the ability of children to jointly reason about agents' costs and rewards when observing short traces of past behaviour \cite{jara2016naive, liu2017ten}.

\subsection{Learning to model deep RL agents}
\label{section:trained-agents}

The previous experiments demonstrate the ToMnet's ability to learn models of simple, algorithmic agents which have full observability. We next considered the ToMnet's ability to learn models for a richer population of agents: those with partial observability and neural network-based policies, trained using deep reinforcement learning. In this section we show how the ToMnet learns how to do inference over the kind of deep RL agent it is observing, and show the specialised predictions it makes as a consequence.

This domain begins to capture the complexity of reasoning about real-world agents. So long as the deep RL agents share some overlap in their tasks, structure, and learning algorithms, we expect that they should exhibit at least some shared behavioural patterns. These patterns should also diverge systematically from each other as the aforementioned factors vary, and individual agents may also exhibit idiosyncratic behaviour as they settle in local minima while optimising their respective policies. There are thus opportunities to learn rich general and agent-specific theories of mind for such populations. Moreover, as the tasks and networks become more complex, hand-crafting a Machine Theory of Mind to parse behaviour based on our human knowledge \citep[e.g.\ ][]{baker2011bayesian, nakahashi2016modeling, Baker2017, lake2017building} becomes increasingly intractable; instead we seek here a path towards machines which learn how to model others' minds autonomously \cite{botvinick2017building}.

We trained three different species of agents on a modified version of the gridworlds, described below in Section~\ref{section:implicit-beliefs}. In brief, agents received maximum reward for reaching a subgoal location first, then consuming a preferred object that differed from agent to agent. Consuming any of the non-subgoal objects terminated the episode. All agents were based on the UNREAL architecture \cite{jaderberg2017reinforcement}, with details given in Appendix~\ref{appendix:agents}. One species of agent (``blind'') was unable to observe the maze state at all, and could only observe its previous action ($a_{t-1}$) and reward ($r_{t-1}$), which it could integrate over time through its LSTM state. The second species had partial observability (``sighted''), but was stateless: these agents could observe the gridworld within a $5\times5$ window centred at their current location, with the rest of the maze shrouded by a binary mask; the agents' policies however were purely reactive, implemented via feedforward networks without any memory. The third species shared the benefits of the other two, being both sighted (with partial observability) and stateful (with an LSTM-based policy).

The ToMnet, on the other hand, observed the behaviour of these agents with full observability of the POMDP state. We constructed a training set for the ToMnet by generating behaviour from 120 trained agents (3 species $\times$ 4 preferred objects $\times$ 10 neural nets trained with different random seeds). We held out a test set of a further 120 trained agents (i.e.\ trained using 10 additional random seeds) for evaluating the trained ToMnet. All results below are shown for the test set, i.e.\ predicting behaviour for novel agents which the ToMnet had never seen before.

\begin{figure}[!t]
\begin{center}
\centerline{\includegraphics[width=\columnwidth]{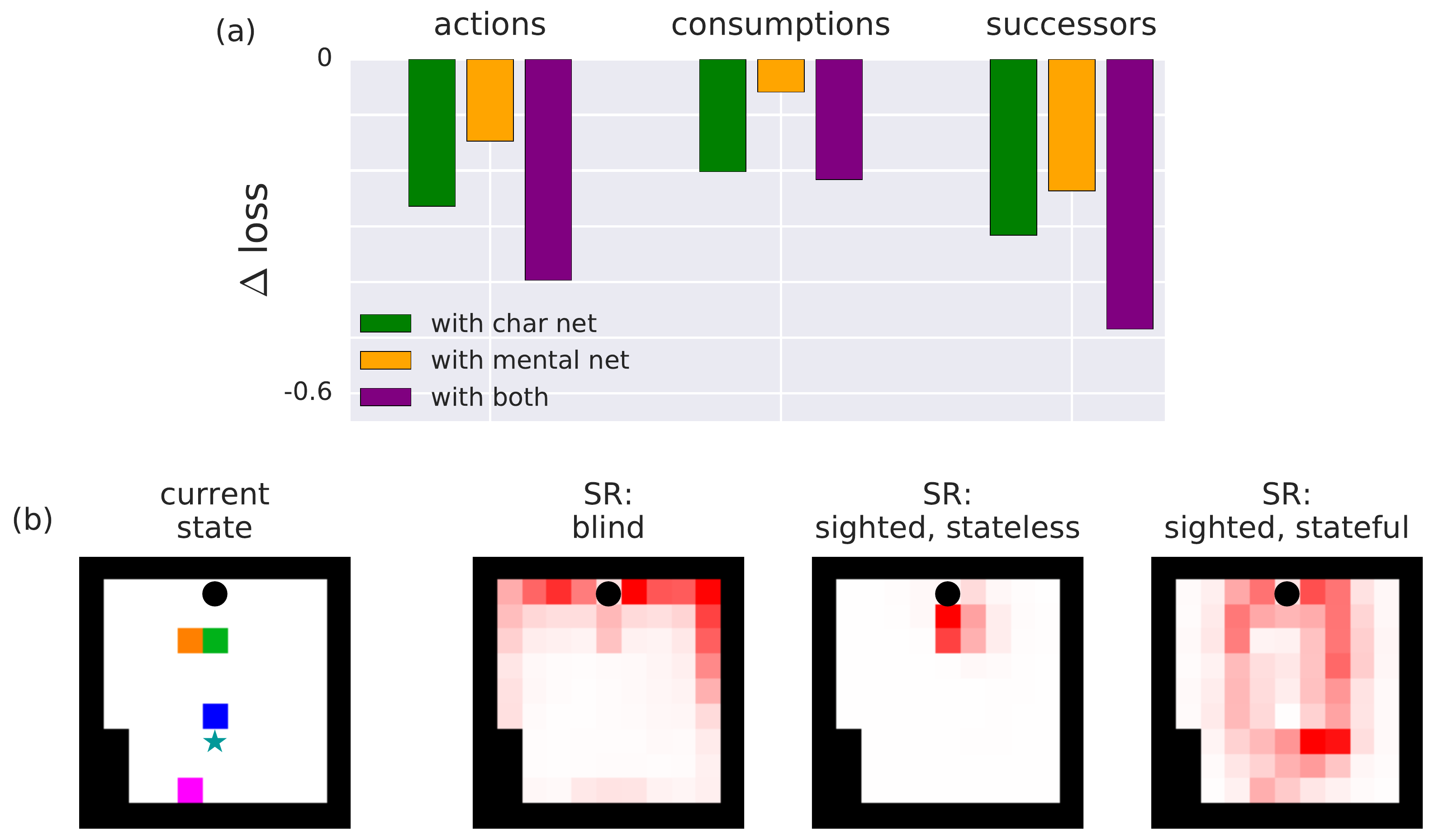}}
	\caption{{\leg Using the ToMnet to characterise trained neural-net agents.} {\leg (a)} Usefulness of ToMnet components for the three behavioural prediction targets, compared with a simple ToMnet with no character nor mental net. Longer bars are better; including both character and mental nets is best. More details are given in Table~\ref{appendix:table:train-vs-test}. {\leg (b)} A ToMnet's prediction of agents' future state occupancy given a query POMDP state at time $t=0$ (left), as per Fig~\ref{fig:goal-driven-1}d. Star denotes the subgoal. The maps on the right are produced after observing behaviour on $\Npast=5$ past POMDPs from a sampled agent of each subspecies (always preferring the pink object). The ToMnet does not know a priori which subspecies each agent belongs to, but infers it from past behaviour.}
\label{fig:trained-agents-1}
\end{center}
\vskip -0.2in
\end{figure}

Unlike previous experiments, these agents' behaviour depended on both their individual characteristics and their state; the ToMnet thus needed both a character net and a mental net to make the best predictions (Fig~\ref{fig:trained-agents-1}a). 

Qualitative evaluations of the ToMnet's predictions show how it learned the expected behaviour of the three species of agents. Fig~\ref{fig:trained-agents-1}b shows the ToMnet's predictions of future state occupancy for the same query state, but given different past observations of how the agent behaves. Without being given the species label, the ToMnet implicitly infers it, and maps out where the agent will go next: blind agents continue until they hit a wall, then turn; sighted but stateless agents consume objects opportunistically; sighted, stateful agents explore the interior and seek out the subgoal. Thus the ToMnet develops general models for the three different species of agents in its world.

\begin{figure}[!t]
\begin{center}
\centerline{\includegraphics[width=\columnwidth]{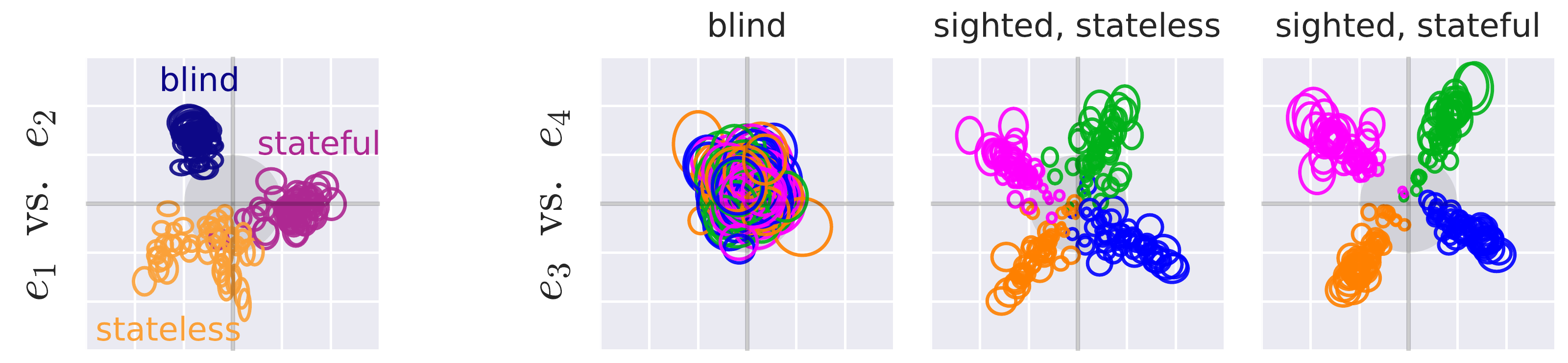}}
	\caption{{\leg Variational character embeddings produced from observations of held-out agents.} This shows how the ToMnet learns a disentangled conceptual space of agent personalities. Left panel shows the first two of four non-degenerate dimensions of $\echar \in \mathbb{R}^8$; right panels show the second two. Ellipses denote the Gaussian covariance (one stdev) of the posteriors $q(\echar|\cdot)$. Left: posteriors coloured by agents' ground-truth species. Right: posteriors coloured by agents' ground-truth preferred objects. The ToMnet uses the first two dimensions of $\echar$ (left panel) to represent which of the three species the agent belongs to, and the next two dimensions (right panels) to represent its preferred object. When the agent is blind, the ToMnet represents the agent's preferred object by the prior, a unit Gaussian. All posteriors collapsed to the prior in the remaining four dimensions.}
\label{fig:trained-agents-2}
\end{center}
\vskip -0.2in
\end{figure}

\begin{figure}[!t]
\begin{center}
\centerline{\includegraphics[width=\columnwidth]{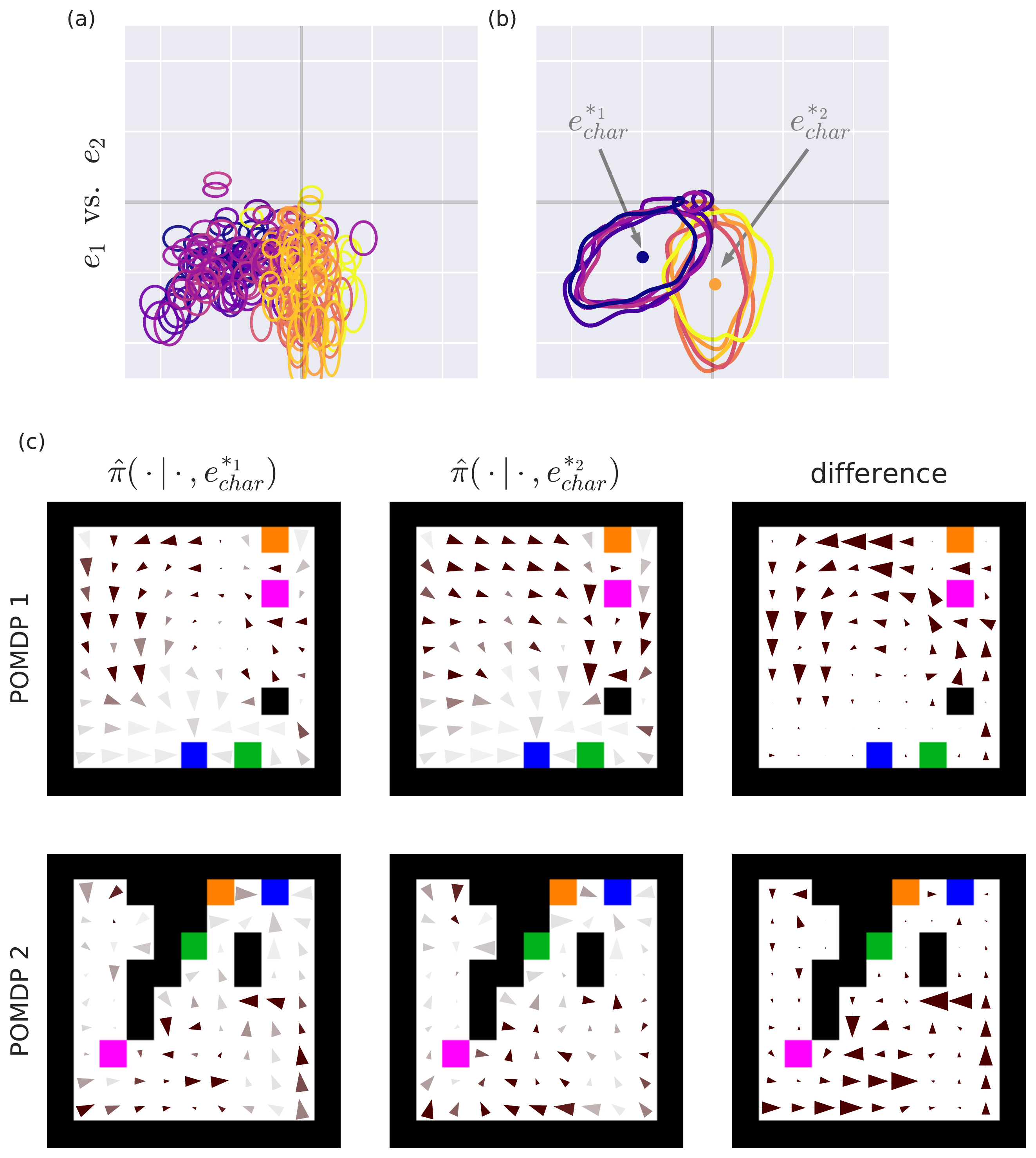}}
	\caption{{\leg The ToMnet discovers two subspecies of sighted/\allowbreak stateless agents.} {\leg (a)} Variational character posteriors, $q(\echar)$, for sighted/\allowbreak stateless agents. Axes show the first two non-degenerate dimensions of $\echar$ (as in Fig~\ref{fig:trained-agents-2}a). Each colour shows the posteriors inferred from a single deep RL agent from the test set, using different behavioural traces. {\leg (b)} Marginal posteriors for the individual agents shown in (a). These are shown as iso-density contours, enclosing 80\% of the total density. Dots show the cluster means. {\leg (c)} Predicted policy differences between agents in the two clusters. Each row shows a different query POMDP. Each panel shows predicted policy for different agent locations, as in Fig~\ref{fig:goal-driven-2}c. Left: ToMnet's prediction for an agent with $\echar$ at the one cluster mean. Middle: at the other cluster mean. Arrows are darker where the two policies differ (higher $D_{JS}$). Right: vector difference between left and middle. Agents in the first cluster explore in an anti-clockwise direction, while agents in the second cluster explore in a clockwise direction.}
\label{fig:trained-agents-3}
\end{center}
\vskip -0.2in
\end{figure}

While we wished to visualise the agent embeddings as in previous experiments, constraining $\echar$ to a 2D space produced poor training performance. With the higher dimensionality required to train the ToMnet on this task (e.g.\ using $\mathbb{R}^8$), we found it difficult to discern any structure in the embedding space. This was likely due to the relatively deep prediction network, and the lack of explicit pressure to compress or disentangle the embeddings. However, the results were dramatically different when we added an explicit bottleneck to this layer, using the Deep Variational Information Bottleneck technique recently developed in \citet{alemi2016deep}. By replacing the character embedding vectors $\echar$ with simple Gaussian posteriors, $q(\echar|\cdot)$, limiting their information content by regularising them towards a unit Gaussian prior, $p(\echar)$, and annealing the respective penalty, $\mathcal{L}_{q} = \beta \, D_{KL}(q||p)$ from $\beta=0$ to $\beta=0.01$ over training, the ToMnet was driven to disentangle the factors of variation in agent personality space (Fig~\ref{fig:trained-agents-2}). Moreover, the ToMnet even discovered substructure amongst the sighted/stateless subspecies that we were not aware of, as it clustered sighted/stateless test agents into two subcategories (Fig~\ref{fig:trained-agents-3}a-b). By contrasting the ToMnet's predictions for these two clusters, the structure it discovers becomes obvious: each sighted/stateless agent explores its world using one of two classic memoryless wall-following algorithms, the {\it right-hand rule} or the {\it left-hand rule} (Fig~\ref{fig:trained-agents-3}c).

\subsection{Acting based on false beliefs}
\label{section:implicit-beliefs}

It has long been argued that a core part of human Theory of Mind is that we recognise that other agents do not base their decisions directly on the state of the world, but rather on an {\it internal representation} of the state of the world \cite{leslie1987pretense, gopnik1988children, wellman1992child, baillargeon2016psychological}. This is usually framed as an understanding that other agents hold {\it beliefs} about the world: they may have knowledge that we do not; they may be ignorant of something that we know; and, most dramatically, they may believe the world to be one way, when we in fact know this to be mistaken. An understanding of this last possibility -- that others can have {\it false beliefs} -- has become the most celebrated indicator of a rich Theory of Mind, and there has been considerable research into how much children, infants, apes, and other species carry this capability \cite{baron1985does, southgate2007action, clayton2007social, call2008does, krupenye2016great, baillargeon2016psychological}.

\begin{figure}[!t]
\begin{center}
\centerline{\includegraphics[width=\columnwidth]{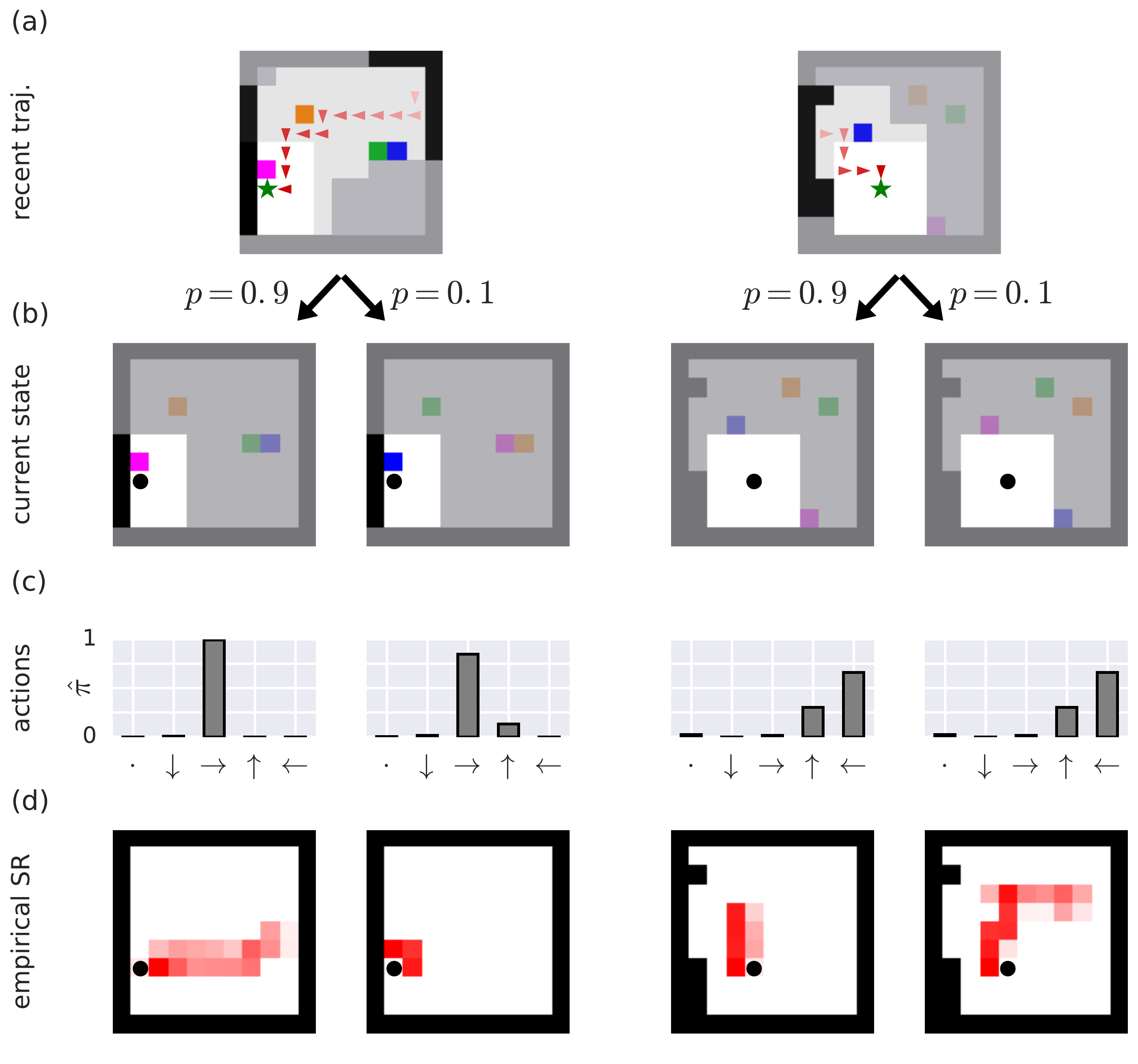}}
	\caption{{\leg Subgoal task, where agents can have false beliefs.} {\leg (a)} Trajectory of an agent (red arrows) as it seeks the subgoal (star). Agent has partial observability: dark grey areas have not been observed; light grey areas have been seen previously, but are not observable at the time of subgoal consumption. {\leg (b)} When the agent consumes the subgoal object, there is a small probability that the other objects will instantaneously swap locations. Left: swap event within the agent's current field of view. Right: outside it. {\leg (c)} Effect of swap on agent's immediate policy. {\leg (d)} Effect of swap on agent's empirical successor representation (average discounted state occupancy over 200 stochastic rollouts). Agent prefers the blue object.}
\label{fig:subgoal-task}
\end{center}
\vskip -0.2in
\end{figure}

Here, we sought to explore whether the ToMnet would also learn that agents may hold false beliefs about the world. To do so, we first needed to generate a set of POMDPs in which agents could indeed hold incorrect information about the world (and act upon this). To create these conditions, we allowed the state of the environment to undergo random changes, sometimes where the agents couldn't see them. In the subgoal maze described above in Section~\ref{section:trained-agents}, we included a low probability ($p=0.1$) state transition when the agent stepped on the subgoal, such that the four other objects would randomly permute their locations instantaneously (Fig~\ref{fig:subgoal-task}a-b). These {\it swap events} were only visible to the agent insofar as the objects' positions were within the agent's current field of view; when the swaps occurred entirely outside its field of view, the agent's internal state and policy at the next time step remained unaffected (policy changes shown in Fig~\ref{fig:subgoal-task}c, right side), a signature of a false belief. As agents were trained to expect these low-probability swap events, they learned to produce corrective behaviour as their policy was rolled out over time (Fig~\ref{fig:subgoal-task}d, right side). While the trained agents were competent at the task, they were not optimal.

In turn, we trained the ToMnet to predict the behaviour of these agents. We initially focused on agents with $5\times5$ fields of view, as in Section~\ref{section:trained-agents}. We trained the ToMnet on rollouts from 40 sighted/stateful agents, each having a preference for one of the four different objects; we tested it on a set of 40 held-out agents. We used the ToMnet model described above in Section~\ref{section:trained-agents}, with $\Npast=4$ past episodes for character inference.

Our goal was to determine whether the ToMnet would learn a general theory of mind that included an element of false beliefs. However, the ToMnet, as described, does not have the capacity to explicitly report agents' (latent) belief states, only the ability to report predictions about the agents' overt behaviour. To proceed, we took inspiration from the literature on human infant and ape Theory of Mind \cite{call2008does, baillargeon2016psychological}. Here, experimenters have often utilised variants of the classic ``Sally-Anne test'' \cite{wimmer1983beliefs, baron1985does} to probe subjects' models of others. In the classic test, the observer watches an agent leave a desired object in one location, only for it to be moved, unseen by the agent. The subject, who sees all, is asked where the agent now believes the object lies. While infants and apes have limited ability to explicitly report such inferences about others' mental states, experimenters have nevertheless been able to measure these subjects' predictions of where the agents will actually go, e.g.\ by measuring anticipatory eye movements, or surprise when agents behave in violation of subjects' expectations \cite{call2008does, krupenye2016great, baillargeon2016psychological}. These experiments have demonstrated that human infants and apes can implicitly model others as holding false beliefs.

\begin{figure}[!t]
\begin{center}
\centerline{\includegraphics[width=\columnwidth]{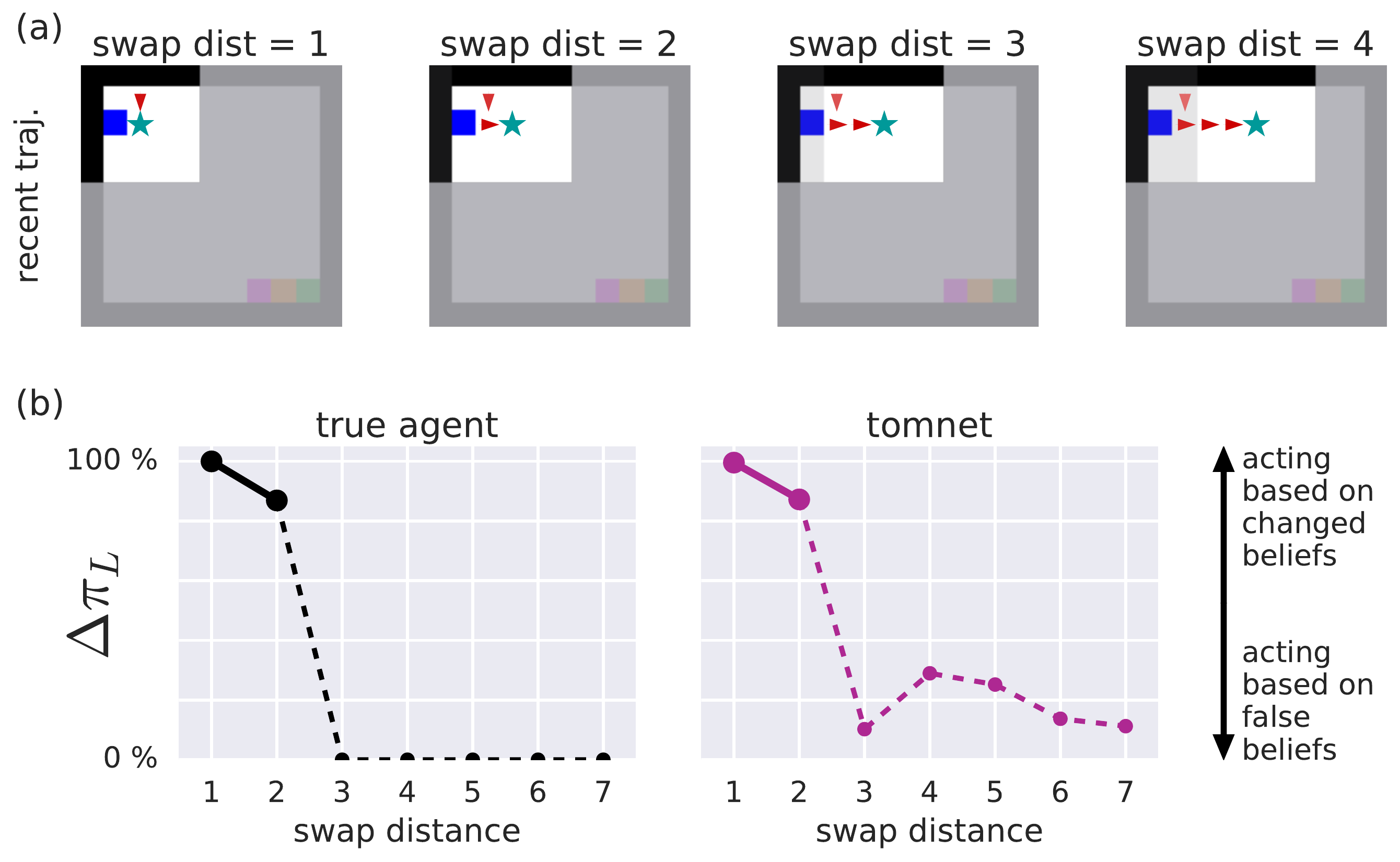}}
	\caption{{\leg Sally-Anne test.} {\leg (a)} We force agents to initially move along a hand-constructed trajectory. Agents have $5\times5$ observability and prefer the blue square object, but must seek the subgoal (star) first. When an agent reaches the subgoal, a swap event may or may not occur. If there is no swap, the optimal action is to go left. By extending the length of the path, the swap event will no longer be visible to the agent. {\leg (b)} Left: effect of a swap event on the agents' true policies, measured as the relative reduction in their probability of moving back towards the original location where they saw the blue object ($\Delta\pi_L = (\pi(a_L|\mathrm{no\;swap}) - \pi(a_L|\mathrm{swap})) / \pi(a_L|\mathrm{no\;swap}) \,\times\,100\%$). If the agent can see that the object has moved from this location (swap dist $\le$ 2), it will not return left. If it cannot see this location, its policy will not change. Right: ToMnet's prediction.}
\label{fig:sally-anne}
\end{center}
\vskip -0.2in
\end{figure}

We used the swap events to construct a gridworld Sally-Anne test. We hand-crafted scenarios where an agent would see its preferred blue object in one location, but would have to move away from it to reach a subgoal before returning to consume it (Fig~\ref{fig:sally-anne}a). During this time, the preferred object might be moved by a swap event, and the agent may or may not see this occur, depending on how far away the subgoal was. We forced the agents along this trajectory (off-policy), and measured how a swap event affected the agent's probability of moving back to the preferred object. As expected, when the swap occurred within the agent's field of view, the agent's likelihood of turning back dropped dramatically; when the swap occurred outside its field of view, the policy was unchanged (Fig~\ref{fig:sally-anne}b, left).

In turn, we presented these demonstration trajectories to the ToMnet (which had seen past behaviour indicating the agent's preferred object). Crucially, the ToMnet was able to observe the {\it entire} POMDP state, and thus was aware of swaps when the agent was not. To perform this task properly, the ToMnet needs to have implicitly learned to separate out what it {\it itself} knows, and what the agent can plausibly know, without relying on a hand-engineered, explicit observation model for the agent. Indeed, the ToMnet predicted the correct behavioural patterns (Fig~\ref{fig:sally-anne}b, right): specifically, the ToMnet predicts that when the world changes far away from an agent, that agent will persist with a policy that is founded on false beliefs about the world.

This test was a hand-crafted scenario. We validated its results by looking at the ToMnet's predictions for how the agents responded to {\it all} swap events in the distribution of POMDPs. We sampled a set of test mazes, and rolled out the agents' policies until they consumed the subgoal, selecting only episodes where the agents had seen their preferred object along the way. At this point, we created a set of counterfactuals: either a swap event occurred, or it didn't.

\begin{figure}[!t]
\begin{center}
\centerline{\includegraphics[width=\columnwidth]{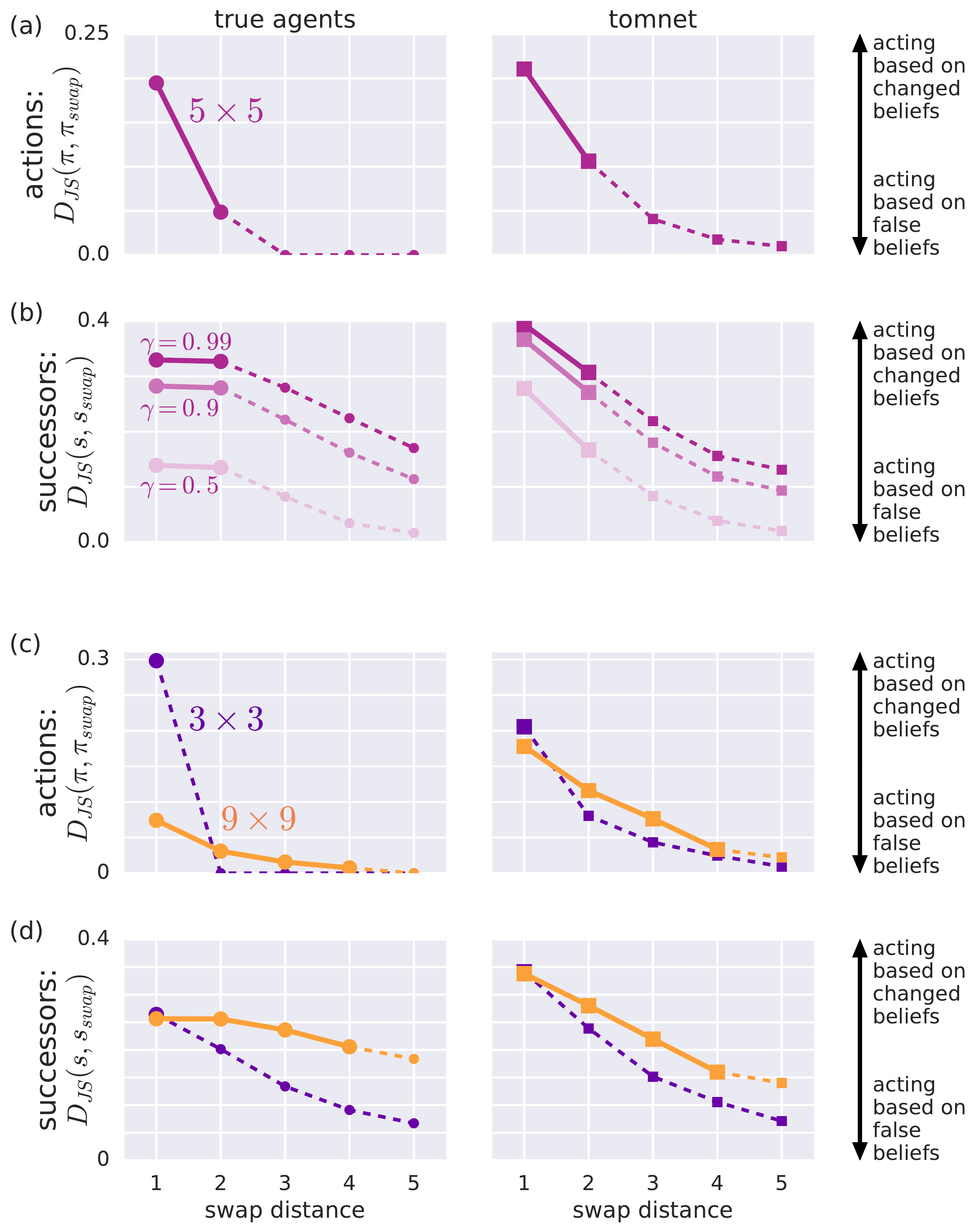}}
	\caption{{\leg Natural Sally-Anne test, using swap events within the distribution of POMDPs.} {\leg (a)} Left: effect of swap events on $5\times5$ agents' next-step policies. Right: ToMnet predictions. {\leg (b)} For SRs of different discount factors ($\gamma$). $D_{JS}$ measured between normalised SRs. {\leg (c)-(d)} As for (a)-(b), but for a ToMnet trained on a range of agents with different fields of view. Showing only $3\times3$ and $9\times9$ results for clarity. For a discussion of why $3\times3$ agents' next-step actions are particularly sensitive to adjacent swap events, see Appendix~\ref{appendix:notes:hypersensitivity}.}
\label{fig:natural-sally-anne}
\end{center}
\vskip -0.2in
\end{figure}

We measured the ground truth for how the swaps would affect the agent's policy, via the average Jensen-Shannon divergence ($D_{JS}$) between the agent's true action probabilities in the no-swap and swap conditions\footnote{For a discussion of why we used the $D_{JS}$ measure, see Appendix~\ref{appendix:notes:JS}.}. As before, the agent's policy often changed when a swap was in view (for these agents, within a 2 block radius), but wouldn't change when the swap was not observable (Fig~\ref{fig:natural-sally-anne}a, left).

The ToMnet learned that the agents' policies were indeed more sensitive to local changes in the POMDP state, but were relatively invariant to changes that occurred out of sight (Fig~\ref{fig:natural-sally-anne}a, right). The ToMnet did not, however, learn a hard observability boundary, and was more liberal in predicting that far-off changes could affect agent policy. The ToMnet also correctly predicted that the swaps would induce corrective behaviour over longer time periods, even when they were not initially visible (Fig~\ref{fig:natural-sally-anne}b).

These patterns were even more pronounced when we trained the ToMnet on mixed populations of agents with different fields of view. In this task, the ToMnet had to infer what each agent could see (from past behaviour alone) in order to predict each agent's behaviour in the future. The ToMnet's predictions reveal an implicit grasp of how different agents' sensory abilities render them differentially vulnerable to acquire false beliefs (Fig~\ref{fig:natural-sally-anne}c-d).

Most surprising of all, we found that the ToMnet learned these statistics even if the ToMnet had never seen swap events during training: the curves in Fig~\ref{fig:natural-sally-anne} were qualitatively identical for the ToMnet under such conditions (Fig~\ref{appendix:fig:natural-sally-anne}).

On the one hand, we were impressed that the ToMnet learns a general theory of mind that incorporates an implicit understanding that agents act based on their own persistent representations of the world, even if they are mistaken. On the other hand, we should not attribute this cognitive ability to a special feature of the ToMnet architecture itself, which is indeed very straightforward. Rather, this work demonstrates that representational Theory of Mind can arise simply by observing competent agents acting in POMDPs.


\subsection{Explicitly inferring belief states}
\label{section:explicit-beliefs}

We have demonstrated that the ToMnet learns that agents can act based on false beliefs. This is limited, though, in that the ToMnet cannot explicitly report what these agents know and don't know about the world. Moreover, it is difficult to extract any beliefs that will not manifest immediately in overt behaviour.

\begin{figure}[!t]
\begin{center}
\centerline{\includegraphics[width=\columnwidth]{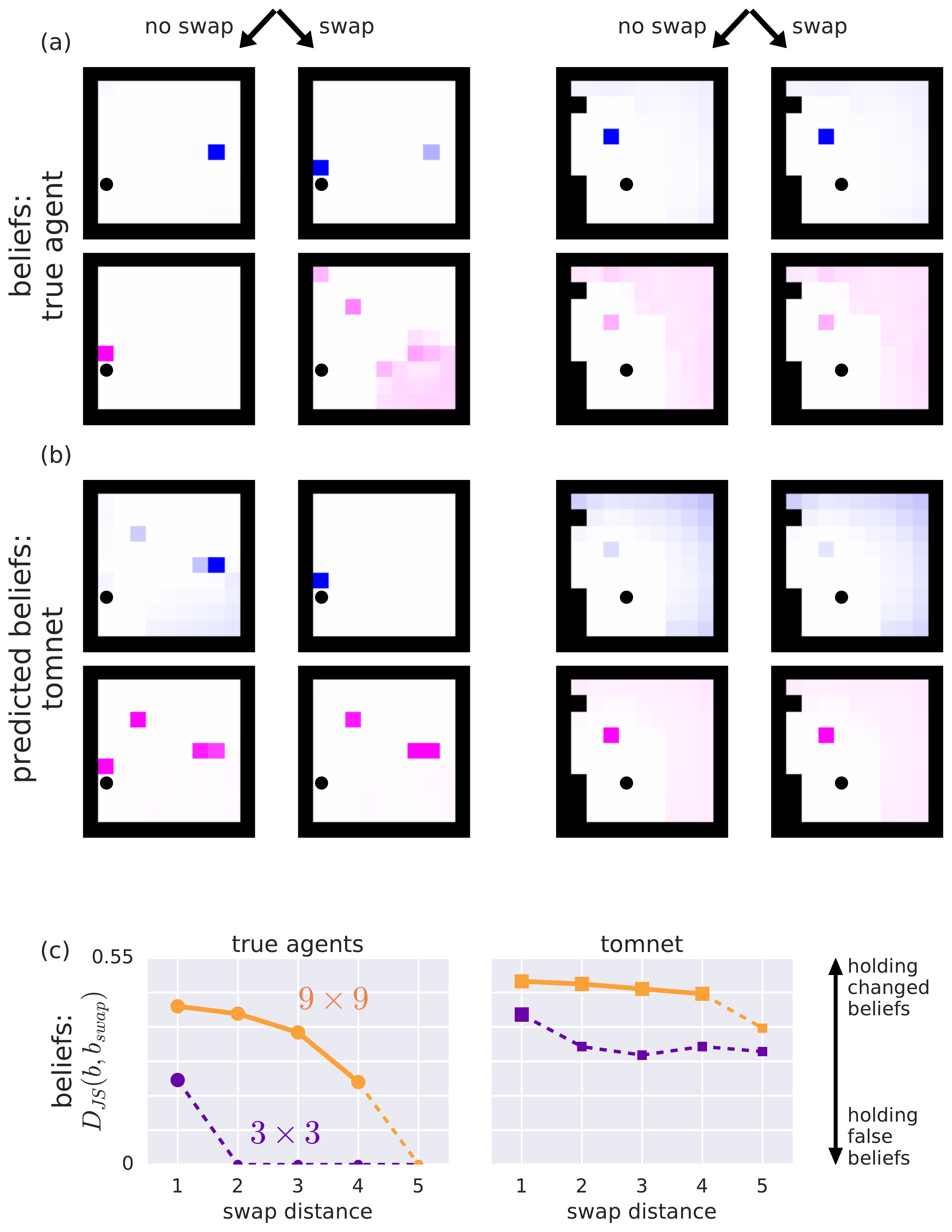}}
	\caption{{\leg Supervised prediction of beliefs.} {\leg (a)} Belief states ($b = p(\mathrm{obj\;location}|x_{0:t_{swap}})$) reported by the agent in the POMDPs shown in Fig~\ref{fig:subgoal-task} at the time of subgoal consumptions. Here we show $b$ for only two objects (rows). In the left two columns, the swap event occurred within the agent's field of view, so the agent's beliefs about the object locations changed given a swap. In the right two columns, the swap event was not within the agent's field of view, so its beliefs about the object locations did not change. {\leg (b)} Predictions $\hat{b}$ made by the ToMnet, given only the trajectory of states and actions. The ToMnet predicts that the observable swap event (left) leads to a change in belief state, whereas the unobservable swap event (right) does not. {\leg (c)} Statistics, as in Fig~\ref{fig:natural-sally-anne}, but for beliefs. The ToMnet captures the longer range over which the $9\times9$ agents can update their beliefs (though again, it infers a soft observability boundary).}
\label{fig:explicit-beliefs}
\end{center}
\vskip -0.2in
\end{figure}

We therefore extended the ToMnet to be able to make declarative statements about agents' beliefs. We achieved this by constructing a supervised dataset of belief states in the sample gridworld. We trained the UNREAL agents to report their beliefs about the locations of the four objects and the subgoal at every time step, alongside their policy. To do this, we added a head to the LSTM that output a posterior over each object's current location on the grid (or whether it was absent). During training, the agents learned to report their best estimate of each object's current location, based on its observations so far during the episode. Example belief states for the query MDP states in Fig~\ref{fig:subgoal-task}a-b are shown in Fig~\ref{fig:explicit-beliefs}a. Note that these reported beliefs are not {\it causal} to the agents' policy; they are just a readout from the same LSTM hidden state that serves the policy.

In turn, we trained a single ToMnet on agents with a range of different fields of view. This ToMnet had to observe only agents' {\it overt} behaviour, and use it to predict, for a given query state, an agent's policy, consumptions, successor representations, and reported beliefs (i.e., what the agent would say about the world if it were asked; e.g.\ Fig~\ref{fig:explicit-beliefs}b).

As shown in Fig~\ref{fig:explicit-beliefs}c, the ToMnet learns agent-specific theories of mind for the different subspecies that grasp the essential differences between their belief-forming tendencies: agents with less visibility of changes in their world are more likely to report false beliefs; and behave according to them too (as in Fig~\ref{fig:explicit-beliefs}c).

\begin{figure}[!t]
\begin{center}
\centerline{\includegraphics[width=\columnwidth]{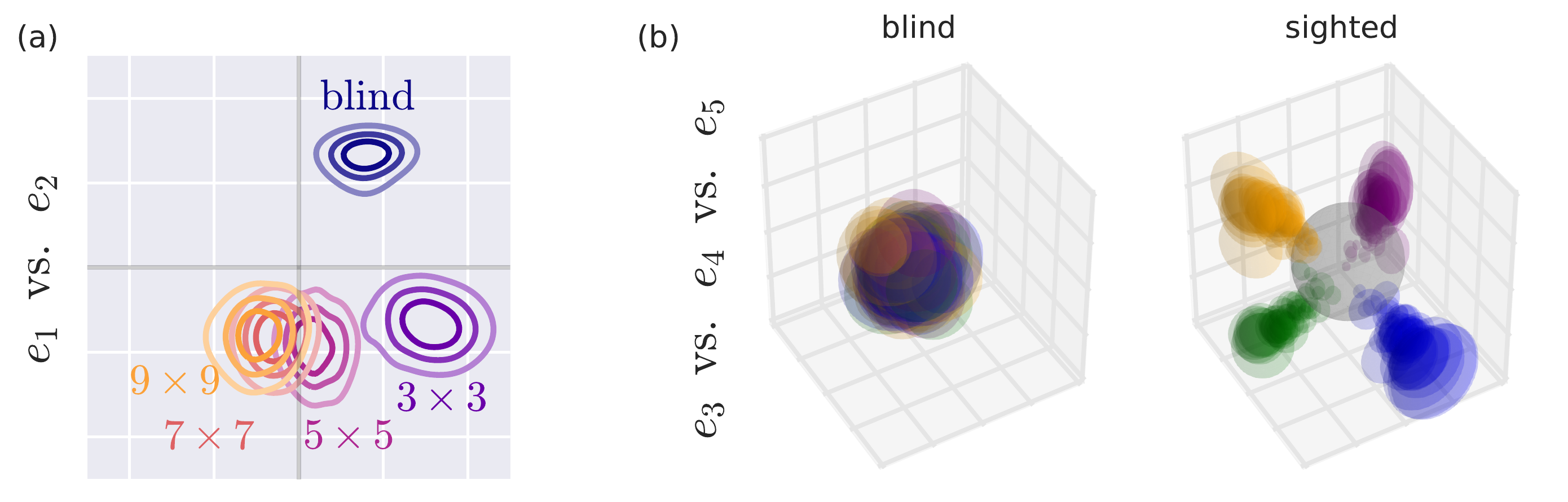}}
	\caption{{\leg Variational character embeddings of agents with different fields of view.} {\leg (a)} First two dimensions of $\echar$ represent field of view. Contours are shown of the marginal posteriors for each agent species. {\leg (b)} Next three dimensions represent preferred objects. Volumes show the approximate marginal posteriors for agents preferring each of the four objects (colours). Blind agents (left) cannot express their preference through their overt behaviour; the ToMnet therefore reverts to the prior. Sighted agents (right) produce embeddings arranged in a roughly tetrahedral arrangement. This same arrangement arises independently of the sighted agents' field of view.}
\label{fig:fov-embeddings}
\end{center}
\vskip -0.2in
\end{figure}

Last of all, we included an additional variational information bottleneck penalty, to encourage low-dimensional abstract embeddings of agent types. As with the agent characterisation in Fig~\ref{fig:trained-agents-1}, the character embeddings of these agents separated along the factors of variation (field of view and preferred object; Fig~\ref{fig:fov-embeddings}). Moreover, these embeddings show the ToMnet's ability to distinguish different agents' visibility: blind and $3\times3$ agents are easily distinguishable, whereas there is little in past behaviour to separate $7\times7$ agents from $9\times9$ agents (or little benefit in making this distinction).

We note that this particular construction of explicit belief inference will likely not scale in its current form. Our method depends on two assumptions that break down in the real world. First, it requires access to others' latent belief states for supervision. We assume here that the ToMnet gets access to these via a rich communication channel; as humans, this channel is likely much sparser. It is an empirical question as to whether the real-world information stream is sufficient to train such an inference network. We do, however, have privileged access to some of our own mental states through meta-cognition; though this data may be biased and noisy, it might be sufficiently rich to learn this task. Second, it is intractable to predict others' belief states about every aspect of the world. As humans, we nevertheless have the capacity to make such predictions about arbitrary variables as the need arises. This may require creative solutions in future work, such as forming abstract embeddings of others' belief states that can be queried.


\section{Discussion}
\label{discussion}

In this paper, we used meta-learning to build a system that learns how to model other agents. We have shown, through a sequence of experiments, how this {\it ToMnet} learns a general model for agents in the training distribution, as well as how to construct an agent-specific model online while observing a new agent's behaviour. The ToMnet can flexibly learn such models over a range of different species of agents, whilst making few assumptions about the generative processes driving these agents' decision making. The ToMnet can also discover abstractions within the space of behaviours.

We note that the experiments we pursued here were simple, and designed to illustrate the core ideas and capabilities of such a system. There is much work to do to scale the ToMnet to richer domains.

First, we have worked entirely within gridworlds, due to the control such environments afford. We look forward to extending these systems to operate within complex 3D visual environments, and within other POMDPs with rich state spaces.

Second, we did not experiment here with limiting the observability of the observer itself. This is clearly an important challenge within real-world social interaction, e.g.\ when we try to determine what someone else knows that we do not. This is, at its heart, an inference problem \cite{Baker2017}; learning to do this robustly is a future challenge for the ToMnet.

Third, there are many other dimensions over which we may wish to characterise agents, such as whether they are animate or inanimate \cite{scholl2000perceptual}, prosocial or adversarial \cite{ullman2009help}, reactive or able to plan \cite{sutton1998reinforcement}. Potentially more interesting is the possibility of using the ToMnet to discover new structure in the behaviour of either natural or artificial populations, i.e.\ as a kind of machine anthropology.

Fourth, a Theory of Mind is important for social beings as it informs our social decision-making. An important step forward for this research is to situate the ToMnet inside artificial agents, who must learn to perform multi-agent tasks.

In pursuing all these directions we anticipate many future needs: to enrich the set of predictions a ToMnet must make; to introduce gentle inductive biases to the ToMnet's generative models of agents' behaviour; and to consider how agents might draw from their own experience and cognition in order to inform their models of others. Addressing these will be necessary for advancing a Machine Theory of Mind that learns the rich capabilities of responsible social beings.


\section*{Acknowledgements}

We'd like to thank the many people who provided feedback on the research and the manuscript, including Marc Lanctot, Jessica Hamrick, Ari Morcos, Agnieszka Grabska-Barwinska, Avraham Ruderman, Christopher Summerfield, Pedro Ortega, Josh Merel, Doug Fritz, Nando de Freitas, Heather Roff, Kevin McKee, and Tina Zhu.


\bibliography{references}
\bibliographystyle{icml2017}


\clearpage
\appendix
\appendixpage
\section{Model description: architectures}
\label{appendix:tomnet}	

Here we describe the precise details of the architectures used in the main text.

We note that we did not optimise our results by tweaking architectures or hyperparameters in any systematic or substantial way. Rather, we simply picked sensible-looking values. We anticipate that better performance could be obtained by improving these decisions, but this is beyond the scope of this work.

\subsection{Common elements.}

{\bf Pre-processing.} Both the character net and the mental state net consume trajectories, which are sequences of observed state/action pairs, $\tau_{ij}^{(obs)} = \{(x_t^{(obs)}, a_t^{(obs)})\}_{t=0}^{T}$, where $i$ is the agent index, and $j$ is the episode index. The observed states in our experiments, $x_t^{(obs)}$, are always tensors of shape $(11 \times 11 \times K)$, where $K$ is the number of feature planes (comprising one feature plane for the walls, one for each object, and one for the agent). The observed actions, $a_t^{(obs)}$, are always vectors of length 5. We combine these data through a {\it spatialisation-concatenation} operation, whereby the actions are tiled over space into a $(11 \times 11 \times 5)$ tensor, and concatenated with the states to form a single tensor of shape $(11 \times 11 \times (K+5))$.

{\bf Training.} All ToMnets were trained with the Adam optimiser, with learning rate $10^{-4}$, using batches of size 16. We trained the ToMnet for 40k minibatches for random agents (Section~\ref{section:random-agents}), and for 2M minibatches otherwise.

\subsection{ToMnet for random agents (Section~\ref{section:random-agents})}
\label{appendix:random-agents}

{\bf Data.} For each species, $S(\alpha)$, we trained a single ToMnet. For each agent, the ToMnet was provided with a variable number of past episodes ($\Npast \sim U\{0, 10\}$), each of length 1 (i.e.\ each trajectory consisting of a single state-action pair). When no past episodes were sampled for a given agent, the character embedding was set to $\echar = 0$.

{\bf Character net.} Each trajectory $\tau_{ij}$ comprises a single state/action pair. We spatialise the action, and concatenate this with the state. This is passed into a 1-layer convnet, with 8 feature planes and ReLU nonlinearity. We then passed the sequence of these (indexed by $j$) into a convolutional LSTM, with the output passed through an average pooling layer, and a fully-connected layer to a 2D embedding space, to produce $\echari$. We obtained similar results with a wide range of different architectures.

{\bf Mental net.} None.

{\bf Prediction net.} In this experiment, we predict only next-step action (i.e.\ policy, $\hat{\pi}$) We spatialise $\echari$, and concatenate with the query state. This is passed to a 2-layer convnet, with 32 feature planes and ReLUs. This is followed by average pooling, then a fully-connected layer to logits in $\mathbb{R}^5$, followed by a softmax.

\subsection{ToMnet for inferring goals (Section~\ref{section:inferring-goals})}
\label{appendix:inferring-goals}

\subsubsection{Experiment 1: single past MDP}

{\bf Data.} Character embedding formed from a single past episode, comprising a full trajectory on a single MDP. Query state is the initial state of a new MDP, so no mental state embedding required.

{\bf Character net.} For the single trajectory $\tau_{i}$ in the past episode, the ToMnet forms the character embedding $\echari$ as follows. We pre-process the data from each time-step by spatialising the actions, $a_t^{(obs)}$, concatenating these with the respective states, $x_t^{(obs)}$, passing through a 5-layer resnet, with 32 channels, ReLU nonlinearities, and batch-norm, followed by average pooling. We pass the results through an LSTM with 64 channels, with a linear output to either a 2-dim or 8-dim $\echari$ (no substantial difference in results).

{\bf Mental net.} None.

{\bf Prediction net.} In this and subsequent experiments, we make three predictions: next-step action, which objects are consumed by the end of the episode, and successor representations. We use a shared torso for these predictions, from which separate heads branch off. For the prediction torso, we spatialise $\echari$, and concatenate with the query state; this is passed into a 5-layer resnet, with 32 channels, ReLU nonlinearities, and batch-norm.

{\bf Action prediction head.} From the torso output: a 1-layer convnet with 32 channels and ReLUs, followed by average pooling, and a fully-connected layer to 5-dim logits, followed by a softmax. This gives the predicted policy, $\hat{\pi}$.

{\bf Consumption prediction head.} From the torso output: a 1-layer convnet with 32 channels and ReLUs, followed by average pooling, and a fully-connected layer to 4-dims, followed by a sigmoid. This gives the respective Bernoulli probabilities that each of the four objects will be consumed by the end of the episode, $\hat{c}$.

{\bf Successor representation prediction head.} From the torso output: a 1-layer convnet with 32 channels and ReLUs, then a 1-layer convnet with 3 channels, followed by a softmax over each channel independently. This gives the predicted normalised SRs for the three discount factors, $\gamma = 0.5, 0.9, 0.99$.

\subsubsection{Experiment 2: many past MDPs, only a single snapshot each}

{\bf Data.} Character embedding is formed from many past episodes ($\Npast \sim U\{0, 10\}$); however, we only use a snapshot of a single time point (i.e.\ a single state/action pair) from each past episode. Query state as for Experiment 1.

{\bf Character net.} For each trajectory $\tau_{ij}$, the character net adds a contribution $\echarij$ to the character embedding $\echari$ as follows. We pre-process the single state/action pair in $\tau_{ij}$ as in Experiment 1, i.e.\ spatialising, concatenating, resnet, average pooling. There is no LSTM as there is only a single state/action pair; instead we use a fully-connected layer to form $\echarij \in \mathbb{R}^2$. These are summed across the $\Npast$  past trajectories to form $\echari$.

{\bf Mental net.} None.

{\bf Prediction net.} As for Experiment 1.

\subsubsection{Experiment 3: greedy agents}

Same as Experiment 1, but with $\Npast \in U\{0, 5\}$.

\subsection{ToMnet for modelling deep RL agents (Section~\ref{section:trained-agents})}
\label{appendix:trained-agents}

{\bf Data.} Character embedding is formed from observing full trajectories of agents on $\Npast = 4$ POMDPs. While the agents have partial observability, potentially with significant parts of the state masked, the observer sees the entire gridworld (albeit without any indication of the field of view of the agent). The current episode is split at a random time (drawn uniformly from $U\{0, T-1\}$ where $T$ is the length of the trajectory). The trajectory prior to the split forms the ``recent trajectory'', and is passed to the mental net. The state at the time of the split is used as the query state.

{\bf Character net.} For each past trajectory $\tau_{ij}$, the character net adds a contribution $\echarij$ to the character embedding $\echari$ via the same architecture as in Experiment 1 described in Appendix~\ref{appendix:inferring-goals} above, with an 8-dim $\echarij$. These are summed to form $\echari$.

{\bf Mental net.} We pre-process each time step's state/action pair in the recent trajectory as follows: we spatialise the action, concatenate with the state, pass through a 5-layer resnet, with 32 channels, ReLU nonlinearities, and batch-norm. The results are fed into a convolutional LSTM with 32 channels. The LSTM output is also a 1-layer convnet with 32 channels, yielding a mental state embedding $\ementali \in \mathbb{R}^{11 \times 11 \times 32}$. When the recent trajectory is empty (i.e.\ the query state is the initial state of the POMDP), $\ementali$ is the zero vector.

{\bf Prediction net.} As in Experiment 1 described in Appendix~\ref{appendix:inferring-goals}. However, the prediction torso begins by spatialising $\echari$ and concatenating it with both $\ementali$ and the query state. Also, as these agents act in gridworlds that include the subgoal object, the consumption prediction head outputs a 5-dim vector.

{\bf DVIB.} For the Deep Variational Information Bottleneck experiments, we altered the architecture by making the character net output a posterior density, $q(\echari)$, rather than a single latent $\echari$; likewise, for the mental net to produce $q(\ementali)$, rather than $\ementali$. We parameterised both densities as Gaussians, with the respective nets outputting the mean and log diagonal of the covariance matrices, as in \citet{kingma2013auto}. For the character net, we achieved this by doubling the dimensionality of the final fully-connected layer; for the mental net, we doubled the number of channels in the final convolutional layer. In both cases, we used fixed, isotropic Gaussian priors. For evaluating predictive performance after the bottleneck, we sampled both $\echar$ and $\emental$, propagating gradients back using the reparameterisation trick. For evaluating the bottleneck cost, we used the analytic KL for $q(\echari)$, and the analytic KL for $q(\ementali)$ conditioned on the sampled value of $\echari$. We scaled the bottleneck costs by $\beta_{char} = \beta_{mental} = \beta$, annealing $\beta$ quadratically from 0 to 0.01 over 500k steps.

\subsection{ToMnet for false beliefs (Sections~\ref{section:implicit-beliefs}--\ref{section:explicit-beliefs})}

The ToMnet architecture was the same as described above in Appendix~\ref{appendix:trained-agents}. The experiments in Section~\ref{section:explicit-beliefs} also included an additional belief prediction head to the prediction net.

{\bf Belief prediction head.} For each object, this head outputs a 122-dim discrete distribution (the predicted belief that the object is in each of the $11 \times 11$ locations on the map, or whether the agent believes the object is absent altogether). From the torso output: a 1-layer convnet with 32 channels and ReLU, branching to (a) another 1-layer convnet with 5 channels for the logits for the predicted beliefs that each object is at the $11\times11$ locations on the map, as well as to (b) a fully-connected layer to 5-dims for the predicted beliefs that each object is absent. We unspatialise and concatenate the outputs of (a) and (b) in each of the 5 channels, and apply a softmax to each channel.

\newpage
\section{Loss function}
\label{appendix:loss}

Here we describe the components of the loss function used for training the ToMnet.

For each agent, $\Agent_i$, we sample past and current trajectories, and form predictions for the query POMDP at time $t$. Each prediction provides a contribution to the loss, described below. We average the respective losses across each of the agents in the minibatch, and give equal weighting to each loss component.

{\bf Action prediction.} The negative log-likelihood of the true action taken by the agent under the predicted policy:
\begin{equation*}
\mathcal{L}_{\mathrm{action}, i} = -\log \hat{\pi}(a_t^{(obs)} | x_t^{(obs)}, \echari, \ementali)
\end{equation*}

{\bf Consumption prediction.} For each object, $k$, the negative log-likelihood that the object is/isn't consumed: 
\begin{equation*}
\mathcal{L}_{\mathrm{consumption}, i} = \sum_k -\log p_{c_k}(c_k | x_t^{(obs)}, \echari, \ementali)
\end{equation*}

{\bf Successor representation prediction.} For each discount factor, $\gamma$, we define the agent's empirical successor representation as the normalised, discounted rollout from time $t$ onwards, i.e.:
\begin{equation*}
    SR_{\gamma}(s) = \frac 1 Z \sum_{\Delta t = 0}^{T - t} \gamma^{\Delta t} I(s_{t + \Delta t} = s)
\end{equation*}
where $Z$ is the normalisation constant such that $\sum_{s} SR_{\gamma}(s) = 1$. The loss here is then the cross-entropy between the predicted successor representation and the empirical one:
\begin{equation*}
    \mathcal{L}_{\mathrm{SR}, i} = \sum_{\gamma} \sum_s -SR_{\gamma}(s) \log \widehat{SR}_{\gamma}(s)
\end{equation*}

{\bf Belief prediction.} The agent's belief states for each object $k$ is a discrete distribution over 122 dims (the $11 \times 11$ locations on the map, plus an additional dimension for an absent object), denoted $b_k(s)$. For each object, $k$, the loss is the cross-entropy between the ToMnet's predicted belief state and the agent's true belief state:
\begin{equation*}
    \mathcal{L}_{\mathrm{belief}, i} = \sum_{k} \sum_s -b_{k}(s) \log \hat{b}_{k}(s)
\end{equation*}

{\bf Deep Varational Information Bottleneck.} In addition to these loss components, where DVIB was used, we included an additional term for the $\beta$-weighted KLs between posteriors and the priors
\begin{align*}
    \mathcal{L}_{DVIB} \;=\;
        &\beta D_{KL}\left( q(\echari) || p(\echar) \right) + \\
        &\beta D_{KL}\left( q(\ementali) || p(\emental) \right)
\end{align*}

\newpage
\section{Gridworld details}
\label{appendix:gridworlds}

The POMDPs $\MDP_j$ were all $11 \times 11$ gridworld mazes. Mazes in Sections~\ref{section:random-agents}--\ref{section:inferring-goals} were sampled with between 0 and 4 random walls; mazes in Sections \ref{section:trained-agents}--\ref{section:explicit-beliefs} were sampled with between 0 and 6 random walls. Walls were defined between two randomly-sampled endpoints, and could be diagonal.

Each $\MDP_j$ contained four terminal objects. These objects could be consumed by the agent walking on top of them. Consuming these objects ended an episode. If no terminal object was consumed after 31 steps (random and algorithmic agents; Sections~\ref{section:random-agents}--\ref{section:inferring-goals}) or 51 steps (deep RL agents; Sections~\ref{section:trained-agents}--\ref{section:explicit-beliefs}), the episodes terminated automatically as a time-out. The sampled walls may trap the agent, and make it impossible for the agent to terminate the episode without timing out.

Deep RL agents (Sections~\ref{section:trained-agents}--\ref{section:explicit-beliefs}) acted in gridworlds that contained an additional subgoal object. Consuming the subgoal did not terminate the episode.

Reward functions for the agents were as follows:

{\bf Random agents (Section~\ref{section:random-agents}.)} No reward function.

{\bf Algorithmic agents (Section~\ref{section:inferring-goals}).} For a given agent, the reward function over the four terminal objects was drawn randomly from a Dirichlet with concentration parameter 0.01. Each agent thus has a sparse preference for one object. Penalty for each move: 0.01. Penalty for walking into a wall: 0.05. Greedy agents' penalty for each move: 0.5. These agents planned their trajectories using value iteration, with a discount factor of 1. When multiple moves of equal value were available, these agents sampled from their best moves stochastically.

{\bf Deep RL agents (Sections~\ref{section:trained-agents}--\ref{section:explicit-beliefs}).} Penalty for each move: 0.005. Penalty for walking into a wall: 0.05. Penalty for ending an episode without consuming a terminal object: 1.

For each deep RL agent species (e.g.\ blind, stateless, $5\times5$, ...), we trained a number of canonical agents which received a reward of 1 for consuming the subgoal, and a reward of 1 for consuming a single preferred terminal object (e.g.\ the blue one). Consuming any other object yielded zero reward (though did terminate the episode). We artifically enlarged this population of trained agents by a factor of four, by inserting permutations into their observation functions, $\omega_i$, that effectively permuted the object channels. For example, when we took a trained blue-object-preferring agent, and inserted a transformation that swapped the third object channel with the first object channel, this agent behaved as a pink-object-preferring agent.

\newpage
\section{Deep RL agent training and architecture}
\label{appendix:agents}

Deep RL agents were based on the UNREAL architecture \cite{jaderberg2017reinforcement}. These were trained with over 100M episode steps, using 16 CPU workers. We used the Adam optimiser with a learning rate of $10^{-5}$, and BPTT, unrolling over the whole episode (50 steps). Policies were regularised with an entropy cost of 0.005 to encourage exploration.

We trained a total of 660 agents, spanning 33 random seeds $\times$ 5 fields of view $\times$ 2 architectures (feedforward/convolutional LSTM) $\times$ 2 depths (4 layer convnet or 2 layer convnet, both with 64 channels). We selected the top 20 agents per condition (out of 33 random seeds), by their average return. We randomly partitioned these sets into 10 training and 10 test agents per condition. With the reward permutations described above in Appendix~\ref{appendix:gridworlds}, this produced 40 training and 40 test agents per condition.

{\bf Observations.} Agents received an observation at each time step of nine $11 \times 11$ feature planes -- indicating, at each location, whether a square was empty, a wall, one of the five total objects, the agent, or currently unobservable.

{\bf Beliefs.} We also trained agents with the auxiliary task of predicting the current locations of all objects in the map. To do this, we included an additional head to the Convolutional LSTMs, in addition to the policy ($\pi_t$) and baseline ($V_t$) heads. This head output a posterior for each object's location in the world, $b_k$ (i.e.\ a set of five 122-dim discrete distributions, over the $11\times11$ maze size, including an additional dimension for a prediction that that the object is absent). For the belief head, we used a 3-layer convnet with 32 channels and ReLU nonlinearities, followed by a softmax. This added a term to the training loss: the cross entropy between the current belief state and the true current world state. The loss for the belief prediction was scaled by an additional hyperparameter, swept over the values 0.5, 2, and 5.

\newpage
\section{Additional results}
\label{appendix:results}

\renewcommand{\thefigure}{A\arabic{figure}}
\setcounter{figure}{0}

\renewcommand{\thetable}{A\arabic{table}}
\setcounter{table}{0}

\begin{table}[!h]
\begin{center}
    \begin{tabular}{ | p{3cm} | c | c |}
	    \multicolumn{1}{ c }{Model} &
	    \multicolumn{1}{ c }{Train agents} &
	    \multicolumn{1}{ c }{Test agents} \\
	    \hline
	    \multicolumn{3}{ l }{} \\
	    \multicolumn{3}{ l }{{\bf Action loss}} \\
	    \hline
	    {\bf none} & 1.14 & 1.12 \\ 
	    \hline \hline
	    {\bf char net} & 0.84 & 0.86 \\
	    \hline
	    + shuffled $\echar$ & 1.61 & 1.62 \\ 
	    \hline \hline
	    {\bf mental net} & 0.83 & 0.98 \\
	    \hline
	    + shuffled $\emental$ & 1.61 & 1.65 \\ 
	    \hline \hline
	    {\bf both} & {\bf 0.72} & {\bf 0.73} \\
	    \hline
	    + shuffled $\echar$ & 1.57 & 1.69 \\ 
	    \hline
	    + shuffled $\emental$ & 1.16 & 1.20 \\ 
	    \hline
	    + shuffled both & 1.99 & 2.02 \\ 
	    \hline

	    \multicolumn{3}{ l }{} \\
	    \multicolumn{3}{ l }{{\bf Consumption loss}} \\
	    \hline
	    {\bf none} & 0.34 & 0.36 \\ 
	    \hline \hline
	    {\bf char net} & 0.19 & 0.16 \\
	    \hline
	    + shuffled $\echar$ & 0.83 & 0.77 \\ 
	    \hline \hline
	    {\bf mental net} & 0.32 & 0.30 \\
	    \hline
	    + shuffled $\emental$ & 0.43 & 0.43 \\ 
	    \hline \hline
	    {\bf both} & {\bf 0.16} & {\bf 0.14} \\
	    \hline
	    + shuffled $\echar$ & 0.82 & 0.78 \\ 
	    \hline
	    + shuffled $\emental$ & 0.23 & 0.23 \\ 
	    \hline
	    + shuffled both & 0.83 & 0.77 \\ 
	    \hline

	    \multicolumn{3}{ l }{} \\
	    \multicolumn{3}{ l }{{\bf Successor loss}} \\
	    \hline
	    {\bf none} & 2.48 & 2.53 \\ 
	    \hline \hline
	    {\bf char net} & 2.23 & 2.21 \\
	    \hline
	    + shuffled $\echar$ & 3.17 & 3.13 \\
	    \hline \hline
	    {\bf mental net} & 2.36 & 2.29 \\
	    \hline
	    + shuffled $\emental$ & 2.92 & 2.80 \\
	    \hline \hline
	    {\bf both} & {\bf 2.16} & {\bf 2.04} \\
	    \hline
	    + shuffled $\echar$ & 3.27 & 3.19 \\
	    \hline
	    + shuffled $\emental$ & 2.45 & 2.33 \\
	    \hline
	    + shuffled both & 3.53 & 3.31 \\
	    \hline
    \end{tabular}
\caption{{\leg Full table of losses for the three predictions in Fig~\ref{fig:trained-agents-1}.} For each prediction, we report the loss obtained by a trained ToMnet that had no character or mental net, had just a character net, just a mental net, or both. For each model, we quantify the importance of the embeddings $\echar$ and $\emental$ by measuring the loss when $\echari$ and $\ementali$ are shuffled within a minibatch. The middle column shows the loss for the ToMnet's predictions on new samples of behaviour from the agents used in the trained set. The right column shows this for agents in the test set.}
\label{appendix:table:train-vs-test}
\end{center}
\end{table}

\begin{figure}[!h]
\begin{center}
\centerline{\includegraphics[width=\columnwidth]{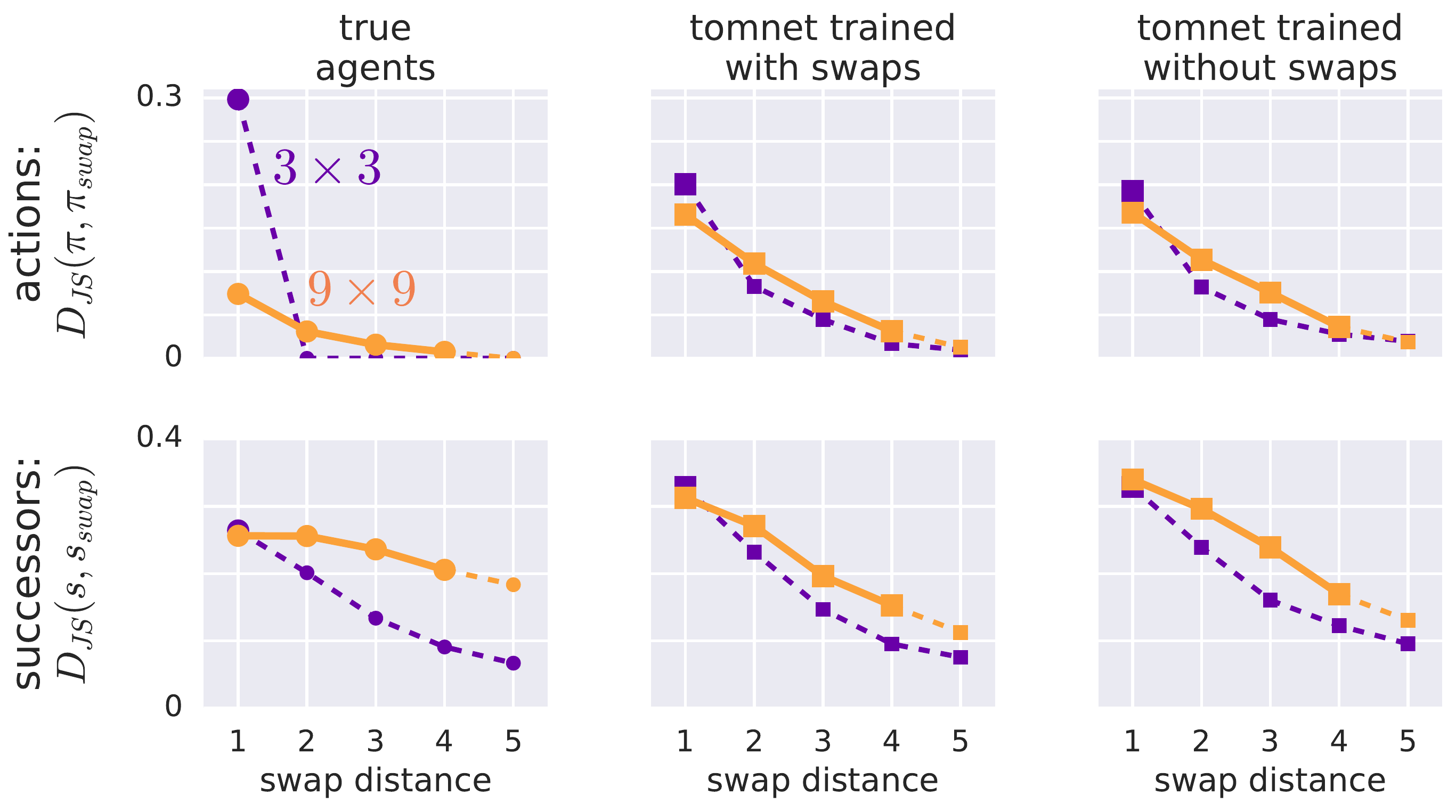}}
\caption{{\leg ToMnet performance on the Natural Sally-Anne test does not depend on the ToMnet observing swap events during training.} The left two columns show the data presented in Fig~\ref{fig:natural-sally-anne} and Fig~\ref{fig:explicit-beliefs}. The rightmost column shows the predictions of the ToMnet when it is trained on data from the same agents, but rolled out on POMDPs where the probability of swap events was $p=0$ instead of $p=0.1$.}
\label{appendix:fig:natural-sally-anne}
\end{center}
\vskip -0.2in
\end{figure}

\newpage
\section{Additional notes}
\label{appendix:notes}

\subsection{Hypersensitivity of $3 \times 3$ agents to swap events with swap distance 1}
\label{appendix:notes:hypersensitivity}

In Fig~\ref{fig:natural-sally-anne}c, the policies of agents with $3 \times 3$ fields of view are seen to be considerably more sensitive to swap events that occur adjacent to the agent than the agents with $9 \times 9$ fields of view. Agents with $5 \times 5$ and $7 \times 7$ had intermediate sensitivities.

We did not perform a systematic analysis of the policy differences between these agents, but we speculate here as to the origin of this phenomenon. As we note in the main text, the agents were competent at their respective tasks, but not optimal. In particular, we noted that agents with larger fields of view were often sluggish to respond behaviourally to swap events. This is evident in the example shown on the left hand side of Fig~\ref{fig:subgoal-task}. Here an agent with a $5 \times 5$ field of view does not respond to the sudden appearance of its preferred blue object above it by immediately moving upwards to consume it; its next-step policy does shift some probability mass to moving upwards, but only a small amount (Fig~\ref{fig:subgoal-task}c). It strongly adjusts its policy on the following step though, producing rollouts that almost always return directly to the object (Fig~\ref{fig:subgoal-task}d). We note that when a swap event occurs immediately next to an agent with a relatively large field of view ($5 \times 5$ and greater), such an agent has the luxury of integrating information about the swap events over multiple timesteps, even if it navigates away from this location. In contrast, agents with $3 \times 3$ fields of view might take a single action that results in the swapped object disappearing altogether from their view. There thus might be greater pressure on these agents during learning to adjust their next-step actions in response to neighbouring swap events.

\subsection{Use of Jensen-Shannon Divergence}
\label{appendix:notes:JS}

In Sections~\ref{section:implicit-beliefs}--\ref{section:explicit-beliefs}, we used the Jensen-Shannon Divergence ($D_{JS}$) to measure the effect of swap events on agents' (and the ToMnet's predicted) behaviour (Figs~\ref{fig:natural-sally-anne}-\ref{fig:explicit-beliefs}). We wanted to use a standard metric for changes to all the predictions (policy, successors, and beliefs), and we found that the symmetry and stability of $D_{JS}$ was most suited to this. We generally got similar results when using the KL-divergence, but we typically found more variance in these estimates: $D_{KL}$ is highly sensitive the one of the distributions assigning little probability mass to one of the outcomes. This was particularly problematic when measuring changes in the successor representations and belief states, which were often very sparse. While it's possible to tame the the KL by adding a little uniform probability mass, this involves an arbitrary hyperparameter which we preferred to just avoid. 


\section{Version history}
\label{appendix:version:history}

\subsubsection*{v1: 21 Feb 2018}

\begin{itemize}
	\item Initial submission to arxiv.
\end{itemize}

\subsubsection*{v2: 12 Mar 2018}

\begin{itemize}
	\item Added missing references to opponent modelling in introduction.
	\item Typographical error in citation \cite{dayan1993improving}.
	\item Figure 3 caption fixes: estimated probabilities $\rightarrow$ action likelihoods; missing descriptor of $\Npast$ for panels (c) and (d).
\end{itemize}

\end{document}